\documentclass[11pt]{article}

\usepackage[final]{acl}

\usepackage{times}
\usepackage{latexsym}

\usepackage[T1]{fontenc}

\usepackage[utf8]{inputenc}

\usepackage{microtype}

\usepackage{inconsolata}

\usepackage{graphicx}

\usepackage{amsmath}
\usepackage{amsthm}
\usepackage{amssymb}    
\usepackage{booktabs}   
\usepackage{colortbl}   
\usepackage{enumitem}   
\usepackage{subcaption} 
\usepackage{algorithm}  
\usepackage{algorithmic}
\usepackage{placeins}   
\newcommand{\fitbox}[1]{\resizebox{\ifdim\width>\linewidth\linewidth\else\width\fi}{!}{#1}}

\newtheorem{theorem}{Theorem}
\newtheorem{lemma}{Lemma}
\newtheorem{corollary}{Corollary}
\newtheorem{proposition}{Proposition}
\theoremstyle{definition}
\newtheorem{definition}{Definition}

\newcommand{\ph}{--}                         

\newcommand{\best}[1]{\textbf{{#1}}}

\title{Closed-Form Fisher-Guided Compression-Rate Allocation for \\ Large Language Models}

\author{
  \normalfont
  Hai-Long Nguyen$^{1}$,
  Sebastiaan B.H.C. Hofstee$^{2}$,
  Trung Thanh Nguyen$^{3}$, \\
  Son Minh Nguyen$^{4}$,
  Faizan Ahmed$^{1}$,
  and Duc V. Le$^{1}$ \\
  $^{1}$University of Twente, The Netherlands \hspace{10pt}
  $^{2}$Rosenxt, The Netherlands \\
  $^{3}$Nagoya University, Japan \hspace{10pt}
  $^{4}$Huawei Technologies Ltd., Finland
}

\begin{document}
\maketitle
\begin{abstract}
Most existing compression methods default to a uniform compression rate across all layers, ignoring the strong heterogeneity of layer importance in large models. Meanwhile, current allocation algorithms are divided into cheap-but-unjustified heuristics and principled-but-costly optimization methods. Thus, no method is both cheap to compute and mathematically tied to the task loss. In this paper, we propose closed-form compression-rate allocation grounded in the empirical Fisher information: a smooth fair rule (DIFW\textsubscript{Fair}) and a knapsack formulation (DTFMW), computed from a single backward pass and provably near-optimal under a second-order approximation of the task loss. Applied on top of low-rank (SVD-based), sparsity, and quantization compressors on LLaMA-2 and LLaMA-3, the proposed allocation matches or beats uniform, searched (Bolaco), and gradient-proportional (FLRC) allocation at the same compression rate in all but one setting. Most notably, Fisher-guided mixed precision rescues 3-bit quantization of LLaMA-3, where uniform allocation collapses in both perplexity and reasoning (402 PPL, near-chance reasoning) while the HAWQ-style instantiation of our loss bound remains usable (19.3 PPL, +10.8 zero-shot points); at 4 bits it still restores near-dense perplexity (9.4 vs.\ 1497, +3.9 points).

\end{abstract}

\section{Introduction}

State-of-the-art AI models deliver remarkable capabilities but typically contain billions of parameters, such as the LLaMA family with 7B to 405B parameters~\cite{touvron2023llama2, dubey2024llama3}, which makes them difficult to deploy on private or resource-constrained hardware; this gap between capability and local deployability motivates efficient model compression research.

Existing compression methods for large Transformer models fall into three main families, each built on a different compression \emph{primitive}, that is, the basic operation applied to each weight matrix.
\emph{Low-rank factorization} decomposes weight matrices into products of smaller ones, from Fisher-Weighted SVD (FWSVD)~\cite{hsulanguage} and its numerically optimized variant TFWSVD~\cite{hua2022numerical} to activation-aware operators such as SVD-LLM~\cite{wang2025svdllm}.
\emph{Sparsity-based pruning} removes weights by magnitude- or activation-based saliency (SparseGPT~\cite{frantar2023sparsegpt}, Wanda~\cite{sun2024wanda}) or deletes entire dimensions (SliceGPT~\cite{ashkboos2024slicegpt}).
\emph{Post-training quantization} lowers weight precision (GPTQ~\cite{frantar2023gptq}, AWQ~\cite{lin2024awq}).
These compressors are highly engineered at the \emph{intra-layer} level, deciding which singular directions, weights, or quantization grids to keep via Fisher weighting, whitening, or Hessian-based updates. At the \emph{inter-layer} level, they assign a uniform compression rate to all layers, ignoring the heterogeneity of layer importance.

Recognizing this, a growing body of work allocates per-layer rates non-uniformly, but existing allocators fall into two camps.
On one side, \emph{cheap heuristics} score layers with hand-crafted signals: activation-outlier ratios in OWL~\cite{yin2024owl}, zeroth-order global scores in ECoFLaP~\cite{sung2024ecoflap}, and gradient-proportional ranks in FLRC~\cite{lu2025flrc}; these run in minutes but offer no formal connection between the score and the task loss.
On the other side, \emph{principled-but-costly methods} optimize an explicit objective: Bayesian search over per-projection ratios in Bolaco~\cite{ji2024bolaco}, gradient-trained truncation in Dobi-SVD~\cite{wang2025dobisvd}, training-time rank adaptation by the TFWSVD authors' group (RankDyna~\cite{hua2023dynamic}; adaptive rank selection~\cite{gao2024adaptive}), evolutionary discovery of allocation functions in DSA~\cite{li2024dsa}, Hessian-trace integer programming in HAWQ~\cite{dong2020hawqv2, yao2021hawqv3}, and closed-form surrogate optimization in MoDeGPT~\cite{lin2025modegpt}; these cost hours of search or training per model, or guarantee only surrogate objectives disconnected from the end-task loss.
Thus, no existing allocation is both cheap to compute and mathematically tied to the task loss.

We close this gap with closed-form compression-rate allocation grounded in the empirical Fisher information: a smooth fair rule (DIFW\textsubscript{Fair}) that assigns each layer a ratio according to its share of task-loss curvature, and a knapsack formulation (DTFMW) that maximizes retained Fisher information under the global budget, where both are computed from a single backward pass and provably near-optimal under a second-order approximation of the task loss.
The contributions of this study are as follows:
\begin{itemize}
    \item We propose two FI-guided allocation methods, a closed-form fair rule (DIFW\textsubscript{Fair}) and a knapsack (DTFMW), and prove a second-order loss bound that explains why the smooth rule is robust while the extremal knapsack needs structural regularization.
    \item We show the allocator is \emph{base-agnostic}: applied on top of the identical base compressor at matched size, the closed-form Fisher rule matches or beats uniform, searched (Bolaco), and proportional (FLRC) allocation in all but one setting across three base compressors, namely low-rank (SVD-LLM), sparsity (Wanda), and quantization (GPTQ), on LLaMA-2-7B and LLaMA-3-8B, at the cost of a single backward pass.
    \item We delimit when each allocation signal wins: external baselines (OWL, MoDeGPT) each specialize to their home primitive, the strongest quantization allocator is the HAWQ-style instantiation of our own bound, and the Fisher rule is the only signal that fails nowhere; a bound-driven sparsity experiment empirically confirms the validity region of the second-order theory. Controlled BERT/GLUE ablations map the three allocation regimes; headline comparisons carry paired multi-seed error bars.
\end{itemize}

\section{Theoretical Analysis}
\label{sec:theory}
 
In this section, we characterize how much task performance degrades as a function of retained Fisher Information (FI) under layer-wise compression, instantiated explicitly for low-rank truncation.
We show that maximizing retained FI minimizes a second-order estimate of the task loss degradation (an upper bound up to a cubic remainder), state precisely when a scalar Fisher rule is optimal, and test that condition empirically --- providing both the justification for and the limits of our FI-guided compression strategies.

\subsection{Notation and Setup}
\label{sec:theory:prelim}

Consider a pre-trained Transformer model with $L$ layers parameterized by $\theta^* = \{\mathbf{W}_1^*, \mathbf{W}_2^*, \ldots, \mathbf{W}_L^*\}$, where $\mathbf{W}_\ell^* \in \mathbb{R}^{m_\ell \times n_\ell}$ denotes the weight matrix of layer $\ell$.
Let $\mathcal{L}(\theta)$ denote the task-specific loss function (e.g., cross-entropy) evaluated on a dataset $\mathcal{D} = \{(x_i, y_i)\}_{i=1}^{N}$, and let $p_\theta(y \mid x)$ denote the model's predictive distribution.
For a model trained to a local minimum, the gradient vanishes: $\nabla_\theta \mathcal{L}(\theta^*) = \mathbf{0}$.

Given any compression scheme that replaces $\mathbf{W}_\ell^*$ with a compressed weight $\widetilde{\mathbf{W}}_\ell$, whether by low-rank truncation, sparsification, or quantization, we define the layer-wise weight perturbation as $\delta \mathbf{W}_\ell = \widetilde{\mathbf{W}}_\ell - \mathbf{W}_\ell^*$ and the vectorized perturbation as $\delta\theta = [\text{vec}(\delta \mathbf{W}_1)^\top, \ldots, \text{vec}(\delta \mathbf{W}_L)^\top]^\top$.
Throughout, $\delta$ denotes a finite change, not an infinitesimal differential; all bounds below are stated in terms of $\delta\theta$ and hold for all three primitives, with low-rank structure entering only in Lemma~\ref{lem:fwsvd_error}.

\subsection{Observed Fisher Information}
\label{sec:theory:fi}

Fisher Information (FI) measures how sensitive a likelihood is to its parameters.
Because the true data distribution is unknown, we work with the \emph{observed} FI, the empirical average over a calibration set $D$; writing the training loss as $\mathcal{L}(d_i; \theta) = -\log f(d_i; \theta)$, it is the mean squared loss gradient:
\begin{equation}
\label{eq:observed_fi}
\widetilde{\mathcal{I}}_\theta = \frac{1}{|D|} \sum_{i=1}^{|D|} \left( \frac{\partial \mathcal{L}(d_i; \theta)}{\partial \theta} \right)^2 .
\end{equation}
Parameters with larger loss gradients are more sensitive to perturbation, so $\widetilde{\mathcal{I}}_\theta$ is a practical, task-aware importance measure available from a single backward pass; it is the quantity our allocators consume (Section~\ref{sec:Method}), previously used to guide pruning and bit assignment~\cite{tu2016ranking, theis2018faster}.

The loss bound derived in the next section requires FI in matrix form and three approximations that make it computable, separable across layers, and expressible on the input side of each weight matrix.
Collecting the per-sample score vectors $\nabla_\theta \log p_{\theta^*}(y \mid x)$ into their expected outer product gives the \textit{Fisher Information Matrix} (FIM) $\mathbf{F}(\theta^*)$ (Appendix~\ref{app:fim}), whose diagonal is estimated by Equation~\eqref{eq:observed_fi}.
Under standard regularity conditions, and when the model is well-specified (i.e., $p_\theta$ is close to the true data distribution), the FIM is equivalent to the Hessian of the negative log-likelihood~\cite{martens2020new}:
\begin{equation}
    \mathbf{F}(\theta^*) \approx \mathbf{H}(\theta^*) = \nabla^2_\theta \mathcal{L}(\theta^*).
    \label{eq:fisher_hessian}
\end{equation}
For layer-wise compression, we adopt the standard \textit{block-diagonal approximation} of the FIM~\cite{martens2015optimizing, singh2020woodfisher}, assuming that cross-layer parameter interactions are negligible:
\begin{equation}
    \mathbf{F}(\theta^*) \approx \mathrm{diag}\left(\mathbf{F}_1, \mathbf{F}_2, \ldots, \mathbf{F}_L\right),
    \label{eq:block_diag}
\end{equation}
where $\mathbf{F}_\ell$ denotes the block of the FIM corresponding to the parameters of layer $\ell$.

Finally, for a linear map $y = \mathbf{W}_\ell x$ each layer block factorizes in Kronecker form into an input second-moment factor $\mathbf{A}_\ell \in \mathbb{R}^{n_\ell \times n_\ell}$ and an output-gradient factor $\mathbf{B}_\ell$ (K-FAC)~\cite{martens2015optimizing, grosse2016kronecker}.
Following FWSVD's row-sum simplification~\cite{hsulanguage}, we absorb the output factor into a global scale, $\mathbf{B}_\ell \propto \mathbf{I}$; the observed FI of Equation~\eqref{eq:observed_fi}, aggregated over output rows, is then a diagonal estimate of $\mathbf{A}_\ell$ up to that scale, which affects no allocation decision.
Throughout, $\mathbf{F}_\ell^{1/2} \triangleq \mathbf{A}_\ell^{1/2}$ denotes this $n_\ell \times n_\ell$ \emph{input-side} factor, which makes every product $\delta\mathbf{W}_\ell \, \mathbf{F}_\ell^{1/2}$ below well-defined.

\subsection{Loss Degradation Bound via Second-Order Taylor Expansion}
\label{sec:theory:taylor}
 
\begin{theorem}[Fisher-Weighted Loss Degradation Bound]
\label{thm:loss_bound}
Let $\theta^*$ be a local minimum of $\mathcal{L}(\theta)$ such that $\nabla_\theta \mathcal{L}(\theta^*) = \mathbf{0}$.
Let $\widetilde{\theta}$ denote the compressed parameters obtained by replacing each layer's weight matrix $\mathbf{W}_\ell^*$ with a compressed weight $\widetilde{\mathbf{W}}_\ell$, produced by any compression primitive.
Under Assumptions 1--4 of Appendix~\ref{sec:theory:assumptions} --- treating the Fisher--Hessian equivalence~\eqref{eq:fisher_hessian}, the block-diagonal structure~\eqref{eq:block_diag}, and the input-side factorization of Section~\ref{sec:theory:fi} as equalities --- the increase in task loss satisfies:
\begin{multline}
    \Delta\mathcal{L} \triangleq \mathcal{L}(\widetilde{\theta}) - \mathcal{L}(\theta^*) \\
    \leq \frac{1}{2} \sum_{\ell=1}^{L} \left\| \delta\mathbf{W}_\ell \, \mathbf{F}_\ell^{1/2} \right\|_F^2 + \mathcal{O}\!\left(\|\delta\theta\|^3\right),
    \label{eq:loss_bound}
\end{multline}
where $\|\cdot\|_F$ denotes the Frobenius norm and $\mathbf{F}_\ell^{1/2}$ is the input-side Fisher factor defined in Section~\ref{sec:theory:fi}.
\end{theorem}

\noindent \textit{Proof in Appendix~\ref{app:proof:thm1}.} At second order,~\eqref{eq:loss_bound} is an equality; the inequality absorbs the cubic remainder (see Limitations and Appendix~\ref{sec:theory:assumptions}).
 
\subsection{Fisher-Guided Allocation for Any Primitive}
\label{sec:theory:general}

Theorem~\ref{thm:loss_bound} bounds the damage of a given perturbation but does not yet choose rates.
Let $s_\ell = m_\ell n_\ell$ denote the size of layer $\ell$, and write $\widetilde{\mathbf{W}}_\ell(r)$ for the weight the primitive returns when layer $\ell$ is compressed to a kept fraction $r \in (0,1]$ of its parameter budget: a rank fraction for low-rank truncation, a weight density for pruning, or a relative bit-width for quantization.
The resulting perturbation $\delta\mathbf{W}_\ell(r)$ defines a \emph{distortion curve}
\begin{equation}
    D_\ell(r) \triangleq \left\| \delta\mathbf{W}_\ell(r) \, \mathbf{F}_\ell^{1/2} \right\|_F^2 ,
    \label{eq:distortion_curve}
\end{equation}
non-increasing in $r$ with $D_\ell(1) = 0$.
Writing $r_\ell$ for the rate assigned to layer $\ell$ and combining~\eqref{eq:distortion_curve} with Theorem~\ref{thm:loss_bound}, spending a global budget $B$ amounts to
\begin{equation}
    \min_{r_1, \ldots, r_L} \; \sum_{\ell=1}^{L} D_\ell(r_\ell)
    \quad \text{s.t.} \quad \sum_{\ell=1}^{L} s_\ell \, r_\ell \leq B .
    \label{eq:general_alloc}
\end{equation}

For discrete primitives each $D_\ell$ is a staircase; the propositions apply to its convex relaxation, and the greedy solver of Section~\ref{sec:Method} handles the discrete case directly.

\begin{proposition}[Optimal allocation for any primitive]
\label{prop:general_alloc}
If every $D_\ell$ is convex, differentiable and non-increasing on $[0,1]$, then a solution of~\eqref{eq:general_alloc} equalizes the marginal damage per parameter across layers: there exists $\lambda \geq 0$ such that $-D_\ell'(r_\ell^*) / s_\ell = \lambda$ for every layer whose $r_\ell^*$ is interior (not pinned at $r = 1$ or at an imposed floor).
\end{proposition}

\begin{proposition}[Fisher sufficiency under distortion homogeneity]
\label{prop:fisher_sufficient}
Suppose the layers share a distortion shape, that is, $D_\ell(r) = \mathrm{FI}_\ell \cdot d(r)$ for a common strictly convex non-increasing $d$, where $\mathrm{FI}_\ell = \|\mathbf{W}_\ell^* \mathbf{F}_\ell^{1/2}\|_F^2$ is the layer's total Fisher energy. Then the optimal rates satisfy
\begin{equation}
    r_\ell^* = \varphi\!\left( \mathrm{FI}_\ell / s_\ell \right)
    \label{eq:fisher_rule}
\end{equation}
for a single increasing function $\varphi$ determined by $d$ and the budget: the optimum depends on a layer \emph{only} through its Fisher density.
\end{proposition}

\noindent \textit{Proofs in Appendices~\ref{app:proof:gen} and~\ref{app:proof:suff}.}

\begin{figure}[t]
\centering
\includegraphics[width=\columnwidth]{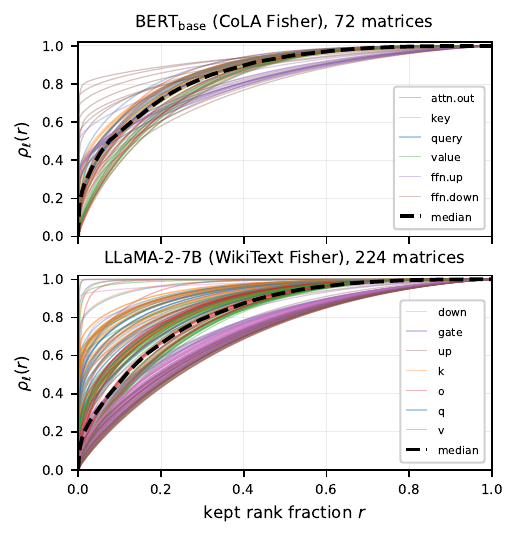}
\caption{\textbf{Testing distortion homogeneity (Proposition~\ref{prop:fisher_sufficient}).} Normalized retained-FI curves $\rho_\ell(r)$ per target matrix (colored by projection type; black dashed: median). Homogeneity holds approximately on BERT\textsubscript{base} (left) but fails on LLaMA-2-7B (right); the Limitations section quantifies the spread and its consequences.}
\label{fig:homogeneity}
\end{figure}

Proposition~\ref{prop:fisher_sufficient} states precisely when a scalar Fisher rule can be optimal.
DIFW\textsubscript{Fair} (Section~\ref{sec:Method}) is a closed-form rule of this monotone family, deliberately keyed to the Fisher \emph{share} $p_W$ rather than the density --- trading exactness under homogeneity for robustness when it fails --- while DTFMW solves~\eqref{eq:general_alloc} directly from the measured spectra.

Empirically, homogeneity roughly holds on BERT\textsubscript{base} but fails on LLaMA-2-7B (Figure~\ref{fig:homogeneity}; consequences in Limitations), so at scale no scalar rule can reproduce the optimum of~\eqref{eq:general_alloc}; this smooth-but-inexact versus exact-but-brittle tension organizes Section~\ref{sec:experiments}.
For low-rank truncation the distortion is the sum of discarded Fisher-weighted singular values, and rank allocation becomes a fractional knapsack solved exactly by greedy selection (Proposition~\ref{prop:optimal_alloc}), which DTFMW implements (Section~\ref{sec:Method}); the bound also extends to classification error (Theorem~\ref{thm:accuracy_bound}), with all statements, proofs, and assumptions in Appendix~\ref{app:proofs}.

\section{Proposed Method}
\label{sec:Method}
We separate \emph{how} each layer is compressed (the base compressor) from \emph{how much} each layer is compressed (the ratio allocation).
Given any layer-wise base compressor and a global compression target \( c \), the proposed methods allocate per-layer budgets from a single diagonal Fisher estimate, obtained with one backward pass over calibration data.
We propose two allocation methods.
\textbf{DIFW\textsubscript{Fair}} is a closed-form rule that assigns each layer a compression ratio according to its share of the network's total FI; the resulting allocation is \emph{smooth}, keeping every ratio close to the global target.
\textbf{DTFMW} maximizes the total retained FI under the parameter budget by solving a knapsack problem over the Fisher-weighted singular-value spectrum; the resulting allocation is \emph{extremal}, concentrating the budget on the most informative directions.

\vspace{5pt}
\noindent \textbf{Granularity.}
Both methods operate at two granularities: \emph{layer}, one ratio per weight matrix, and \emph{block}, one shared ratio for all projections of a Transformer block.
Block granularity averages Fisher noise over the members of a block and caps the spread of per-layer ratios; as shown in Section~\ref{sec:experiments}, it is the more robust choice on large models.

\subsection{Dynamic Inter-layer Fisher Information Weighted Compression (DIFW\textsubscript{Fair})}
Given a target global compression ratio \( c \), DIFW\textsubscript{Fair} assigns layer-specific ratios by computing each layer's contribution to the total network FI.
For each weight matrix \( W \in Q \) (the set of all target matrices, \( W \in \mathbb{R}^{m \times n} \)), we sum the observed FI over its entries, \( \widetilde{\mathcal{I}_\theta}_{W} = \sum_{i,j} \widetilde{\mathcal{I}_\theta}_{W_{ij}} \), and normalize by the network total \( \widetilde{\mathcal{I}}_{\text{Total}} = \sum_{W' \in Q} \widetilde{\mathcal{I}_\theta}_{W'} \) to obtain the layer's FI share:
\begin{equation}
    p_W = \widetilde{\mathcal{I}_\theta}_{W} \, / \, \widetilde{\mathcal{I}}_{\text{Total}}, \qquad \sum_{W \in Q} p_W = 1.
\end{equation}
Each layer then receives the compression ratio \( c_W = \alpha (1 - p_W) \), where \( \alpha \) is a scalar shared across all \( W \in Q \): a higher \( p_W \) (i.e., higher layer importance) leads to a smaller compression ratio.
Throughout, \( c_W \geq 1 \) denotes a layer's compression ratio and its reciprocal \( y_W = 1/c_W \in (0, 1] \) the layer's kept fraction, so a larger \( c_W \) means stronger compression.
DIFW\textsubscript{Fair} chooses \( \alpha \) such that the \emph{average} ratio across layers equals the global target \( c \), by minimizing the loss:
\begin{equation}
\label{eq:fairinterlayer}
    \mathcal{L}_{\text{Fair}}(\alpha) = \left( c - \frac{1}{|Q|} \sum_{W \in Q} \alpha (1 - p_W) \right)^2.
\end{equation}
Given that \( \sum_{W \in Q} p_W = 1 \) and \( 0 \leq p_W \leq 1 \), minimizing Equation~\ref{eq:fairinterlayer} yields the closed-form solution:
\begin{equation}
    \alpha = \frac{c}{1 - \frac{1}{|Q|}}.
\end{equation}
Although \( \alpha \) itself is independent of \( p_W \), the resulting per-layer compression ratio \( \alpha (1 - p_W) \) still reflects the relative importance of each layer through its FI contribution.
A size-weighted variant (DIFW\textsubscript{Total}), which matches the total parameter count exactly rather than the average ratio, is described in Appendix~\ref{sec:appendix-variants}; it performs on par with DIFW\textsubscript{Fair} in our ablations.

For a low-rank base compressor, the ratio \( c_W = m_W n_W / \left( r_W (m_W + n_W) \right) \) determines the retained rank:
\begin{equation}
    r_W = \frac{m_W n_W}{c_W (m_W + n_W)},
    \label{eq:calculate-rank-k-from-cw}
\end{equation}
so each \( W \) is factored into \( m_W \times r_W \) and \( r_W \times n_W \) matrices.

\subsection{Dynamic Total Fisher Information Maximizing Weighted Compression (DTFMW)}
\label{sec:method:dtfmw}
While DIFW\textsubscript{Fair} distributes the budget proportionally, DTFMW allocates it \emph{optimally} with respect to the loss-degradation bound of Section~\ref{sec:theory}, by maximizing the total retained FI.

\vspace{5pt}
\noindent \textbf{Setup.}
For a layer \( W \in \mathbb{R}^{m_W \times n_W} \) with a diagonal Fisher estimate \( \hat{F}_W \) over its input dimensions, FWSVD computes the SVD of the reweighted matrix \( W \hat{F}_W^{1/2} \), yielding Fisher-weighted singular values \( \sigma_{W,1} \geq \sigma_{W,2} \geq \cdots \geq 0 \).
By Lemma~\ref{lem:fwsvd_error}, retaining rank \( r_W \) preserves the FI value
\begin{equation}
    V_W(r_W) = \sum_{i=1}^{r_W} \sigma_{W,i}^2 ,
    \label{eq:dtfmw-value}
\end{equation}
and discards \( \sum_{i > r_W} \sigma_{W,i}^2 \), which directly upper-bounds the layer's contribution to the task-loss increase.

\vspace{5pt}
\noindent \textbf{Problem.}
Retaining rank \( r_W \) costs \( r_W (m_W + n_W) \) parameters.
With total size \( S_{\text{total}} = \sum_{W} m_W n_W \) and global target \( c \), DTFMW solves:
\begin{align}
    \max_{\{r_W\}} \quad & \sum_{W \in Q} V_W(r_W) \label{eq:dtfmw-knapsack} \\
    \text{s.t.} \quad & \sum_{W \in Q} r_W (m_W + n_W) \leq \frac{S_{\text{total}}}{c}, \nonumber \\
    & r_W \in \{ r_{\min}, \ldots, \min(m_W, n_W) \} \quad \forall\, W \in Q. \nonumber
\end{align}
This is exactly the allocation problem of Proposition~\ref{prop:optimal_alloc}, with the budget measured in parameters rather than ranks and with a rank floor \( r_{\min} \) (discussed below).

\vspace{3pt}
\noindent \textbf{Exact greedy solution.}
Raising \( r_W \) from \( i-1 \) to \( i \) adds value \( \sigma_{W,i}^2 \) at fixed cost \( m_W + n_W \); since the \( \sigma_{W,i}^2 \) are non-increasing, every \( V_W \) is concave and greedily buying the densest remaining increment is provably optimal for the continuous relaxation (a fractional knapsack~\cite{korte2012combinatorial}), coinciding with the top-\(K\) selection of Proposition~\ref{prop:optimal_alloc}; the integer-case gap is at most one increment per layer, negligible at LLM budgets.
Algorithm~\ref{alg:dtfmw} (Appendix~\ref{app:proofs}) implements the greedy with a max-heap: with \( N = \sum_W \min(m_W, n_W) \) candidate increments it runs in \( \mathcal{O}(N \log N) \) time, i.e., seconds even for billion-parameter models, since each spectrum is computed once per layer.

\vspace{5pt}
\noindent \textbf{Block granularity.}
At block granularity, all members of a Transformer block \( b \) share one kept fraction \( y_b \in (0, 1] \).
We precompute per-block value and cost curves
\begin{equation}
\begin{aligned}
    V_b(y) &= \sum_{W \in b} V_W\!\left(r_W(y)\right), \\
    C_b(y) &= \sum_{W \in b} r_W(y)\, (m_W + n_W),
\end{aligned}
\end{equation}
with \( r_W(y) = \left\lfloor y \, m_W n_W / (m_W + n_W) \right\rceil \), on a discrete grid of \( y \), and run the same greedy over block-level increments ranked by \( \Delta V_b / \Delta C_b \).
Sharing one fraction per block bounds the ratio spread inside each block; this is the variant used in our large-model experiments.

\begin{table*}[t]
\centering
\caption{\textbf{Allocation battle on the SVD-based compressor} (SVD-LLM whitening operator), WikiText-2 PPL\(\downarrow\), mean {\tiny$\pm$} std over 5 seeds (Fisher + calibration varied); all rows size-matched to within 2.5\% (rank-rounding residual quantified in Appendix~\ref{sec:appendix:fullgrid}). Cells: PPL {\tiny$\pm$std} / 7-task zero-shot accuracy; best per column in bold by PPL; column headers give the parameters remaining at each kept ratio; ``--'' = zero-shot not evaluated for collapsed models.}
\label{tab:svd-battle}
\setlength{\tabcolsep}{4pt}
\resizebox{1.0\textwidth}{!}
{
\begin{tabular}{lrrrrrr}
\toprule
& \multicolumn{3}{c}{\textbf{LLaMA-2, kept}} & \multicolumn{3}{c}{\textbf{LLaMA-3, kept}} \\
\cmidrule(lr){2-4} \cmidrule(lr){5-7}
\textbf{Allocation} & \textbf{0.4} (2.9B) & \textbf{0.6} (4.1B) & \textbf{0.8} (5.4B) & \textbf{0.4} (3.8B) & \textbf{0.6} (5.2B) & \textbf{0.8} (6.6B) \\
\midrule
\multicolumn{7}{l}{\emph{Baselines}} \\
uniform & 81.52 {\tiny$\pm$2.10} / 30.2 & 16.05 {\tiny$\pm$0.15} / 35.2 & 8.37 {\tiny$\pm$0.04} / 42.3 & 577.8 {\tiny$\pm$58.1} / 29.7 & 73.02 {\tiny$\pm$2.20} / 30.9 & 15.49 {\tiny$\pm$0.08} / 42.8 \\
Bolaco~\cite{ji2024bolaco} & 87.59 {\tiny$\pm$1.52} / 30.4 & 17.88 {\tiny$\pm$0.14} / 35.3 & 8.43 {\tiny$\pm$0.03} / 45.6 & 665.5 {\tiny$\pm$52.8} / 30.0 & \best{64.15 {\tiny$\pm$2.49} / 33.5} & 16.71 {\tiny$\pm$0.09} / 46.2 \\
FLRC~\cite{lu2025flrc} & 147.9 {\tiny$\pm$5.3} / 30.7 & 35.84 {\tiny$\pm$0.43} / 32.4 & 15.23 {\tiny$\pm$0.12} / 37.0 & 2239 {\tiny$\pm$216} / 30.3 & 191.2 {\tiny$\pm$8.4} / 30.5 & 31.92 {\tiny$\pm$0.78} / 35.9 \\
BI (MoDeGPT)~\cite{lin2025modegpt} & 83.85 {\tiny$\pm$2.38} / 30.2 & 16.50 {\tiny$\pm$0.22} / 35.4 & 8.46 {\tiny$\pm$0.04} / 43.0 & 582.2 {\tiny$\pm$59.9} / 29.8 & 71.03 {\tiny$\pm$2.61} / 31.7 & 15.49 {\tiny$\pm$0.09} / 43.5 \\
\midrule
\multicolumn{7}{l}{\emph{Ours (Fisher-guided)}} \\
DIFW\textsubscript{Fair} (L) & 78.51 {\tiny$\pm$2.57} / 30.4 & \best{15.77 {\tiny$\pm$0.17} / 35.4} & \best{8.33 {\tiny$\pm$0.04} / 42.5} & 571.4 {\tiny$\pm$55.0} / 29.7 & 71.66 {\tiny$\pm$2.15} / 31.2 & \best{15.32 {\tiny$\pm$0.06} / 43.0} \\
DIFW\textsubscript{Fair} (B) & \best{72.58 {\tiny$\pm$1.50} / 30.6} & 15.78 {\tiny$\pm$0.16} / 35.8 & 8.46 {\tiny$\pm$0.06} / 42.6 & \best{553.7 {\tiny$\pm$48.4} / 29.7} & 71.34 {\tiny$\pm$2.13} / 30.8 & 15.33 {\tiny$\pm$0.13} / 43.1 \\
DTFMW (L) & \multicolumn{3}{c}{collapse (non-finite)} & 6629 {\tiny$\pm$307} / -- & 20200 {\tiny$\pm$11814} / -- & 12063 {\tiny$\pm$4705} / -- \\
DTFMW (B) & 3602 {\tiny$\pm$354} / 29.5 & 6245 {\tiny$\pm$2285} / 29.1 & 56.80 {\tiny$\pm$22.75} / 32.3 & 1277 {\tiny$\pm$107} / 29.5 & 407.5 {\tiny$\pm$46.6} / 29.7 & 30.83 {\tiny$\pm$1.30} / 35.7 \\
\bottomrule
\end{tabular}
}
\end{table*}

\vspace{5pt}
\noindent \textbf{Rank floor as structural regularization.}
The optimum of Equation~\ref{eq:dtfmw-knapsack} is extremal: layers with a flat Fisher spectrum receive almost no rank.
On billion-parameter models this ``bang-bang'' allocation can push individual layers into a regime where the second-order expansion behind Theorem~\ref{thm:loss_bound} no longer holds (the fourth assumption in Appendix~\ref{sec:theory:assumptions}), and the model collapses even though the predicted bound is small.
The floor \( r_{\min} \) (a minimum kept density for sparsity budgets) bounds each layer's perturbation and restores stability; block granularity acts in the same direction (studied in Section~\ref{sec:experiments}).

\vspace{5pt}
\noindent \textbf{Beyond low-rank budgets.}
Both allocators output a per-layer kept fraction \( y_W = 1/c_W \) that any budgeted compressor can consume: a retained rank, a kept weight density, or an integer bit-width; for DTFMW under non-low-rank primitives, the spectrum in Equation~\ref{eq:dtfmw-value} is replaced by the primitive's per-weight saliency curve, keeping the same greedy solver.

\section{Experiments}
\begin{table*}[t]
\centering
\caption{\textbf{Allocation battle on Wanda (sparsity)}, one-shot, WikiText-2 PPL\(\downarrow\), mean {\tiny$\pm$} std over 5 seeds. Cells: PPL {\tiny$\pm$std} / ZS; best per column in bold by PPL. DTFMW floored at \(0.5\cdot\)kept; BDS = the bound-driven knapsack of Finding (v), which tests the validity region of Theorem~\ref{thm:loss_bound}.}
\label{tab:wanda-battle}
\setlength{\tabcolsep}{4pt}
\resizebox{1.0\textwidth}{!}
{
\begin{tabular}{lrrrrrr}
\toprule
& \multicolumn{3}{c}{\textbf{LLaMA-2, kept}} & \multicolumn{3}{c}{\textbf{LLaMA-3, kept}} \\
\cmidrule(lr){2-4} \cmidrule(lr){5-7}
\textbf{Allocation} & \textbf{0.4} (2.9B) & \textbf{0.6} (4.1B) & \textbf{0.8} (5.4B) & \textbf{0.4} (3.8B) & \textbf{0.6} (5.2B) & \textbf{0.8} (6.6B) \\
\midrule
\multicolumn{7}{l}{\emph{Baselines}} \\
uniform & 10.80 {\tiny$\pm$0.07} / 46.3 & 6.07 {\tiny$\pm$0.01} / 54.5 & \best{5.59 {\tiny$\pm$0.00} / 55.8} & 28.46 {\tiny$\pm$0.20} / 44.8 & 9.39 {\tiny$\pm$0.02} / 57.4 & 8.31 {\tiny$\pm$0.01} / 60.0 \\
Bolaco~\cite{ji2024bolaco} & 125.3 {\tiny$\pm$1.6} / 32.6 & 15.85 {\tiny$\pm$0.11} / 44.7 & 7.58 {\tiny$\pm$0.01} / 50.2 & 81.88 {\tiny$\pm$1.17} / 35.5 & 14.26 {\tiny$\pm$0.04} / 51.9 & 9.88 {\tiny$\pm$0.02} / 56.8 \\
FLRC~\cite{lu2025flrc} & 58.29 {\tiny$\pm$0.83} / 33.0 & 13.30 {\tiny$\pm$0.03} / 41.8 & 8.48 {\tiny$\pm$0.01} / 48.2 & 455.9 {\tiny$\pm$25.4} / 30.4 & 42.79 {\tiny$\pm$0.11} / 42.1 & 13.26 {\tiny$\pm$0.01} / 52.6 \\
OWL~\cite{yin2024owl} & \best{9.19 {\tiny$\pm$0.03} / 48.3} & 6.10 {\tiny$\pm$0.01} / 54.6 & 5.59 {\tiny$\pm$0.00} / 55.3 & \best{22.70 {\tiny$\pm$0.15} / 47.2} & 9.73 {\tiny$\pm$0.02} / 57.5 & 8.39 {\tiny$\pm$0.00} / 60.1 \\
BI (MoDeGPT)~\cite{lin2025modegpt} & 11.61 {\tiny$\pm$0.21} / 46.2 & 6.11 {\tiny$\pm$0.00} / 54.4 & 5.59 {\tiny$\pm$0.00} / 55.5 & 30.02 {\tiny$\pm$0.15} / 44.6 & 9.73 {\tiny$\pm$0.02} / 57.7 & 8.36 {\tiny$\pm$0.01} / 59.7 \\
\midrule
\multicolumn{7}{l}{\emph{Ours (Fisher-guided)}} \\
DIFW\textsubscript{Fair} (L) & 10.51 {\tiny$\pm$0.10} / 46.7 & \best{6.06 {\tiny$\pm$0.01} / 54.4} & 5.59 {\tiny$\pm$0.00} / 55.6 & 28.32 {\tiny$\pm$0.64} / 44.6 & \best{9.38 {\tiny$\pm$0.00} / 57.3} & 8.32 {\tiny$\pm$0.00} / 59.9 \\
DIFW\textsubscript{Fair} (B) & 10.42 {\tiny$\pm$0.11} / 47.0 & 6.10 {\tiny$\pm$0.01} / 54.3 & 5.60 {\tiny$\pm$0.00} / 55.4 & 28.60 {\tiny$\pm$0.40} / 44.6 & 9.42 {\tiny$\pm$0.01} / 57.5 & 8.32 {\tiny$\pm$0.01} / 59.8 \\
DTFMW (L, floored) & 158.2 {\tiny$\pm$7.5} / 30.4 & 8.95 {\tiny$\pm$0.56} / 49.5 & 6.06 {\tiny$\pm$0.01} / 54.4 & 184.7 {\tiny$\pm$5.6} / 31.4 & 15.91 {\tiny$\pm$0.18} / 50.6 & 9.01 {\tiny$\pm$0.04} / 58.2 \\
DTFMW (B, floored) & 89.55 {\tiny$\pm$51.27} / 31.8 & 6.81 {\tiny$\pm$0.09} / 52.8 & 5.59 {\tiny$\pm$0.00} / 55.5 & 57.97 {\tiny$\pm$2.16} / 35.4 & 10.82 {\tiny$\pm$0.12} / 53.9 & \best{8.25 {\tiny$\pm$0.01} / 59.8} \\
BDS (bound-driven, ours) & 52.57 {\tiny$\pm$20.66} / 38.1 & 6.89 {\tiny$\pm$0.13} / 51.6 & 5.60 {\tiny$\pm$0.01} / 54.9 & 65.71 {\tiny$\pm$3.10} / 38.5 & 10.16 {\tiny$\pm$0.02} / 56.4 & 8.36 {\tiny$\pm$0.01} / 59.7 \\
\bottomrule
\end{tabular}
}
\end{table*}

\label{sec:experiments}
\subsection{Experimental Setup}
\noindent \textbf{Models and base compressors.}
Our main experiments compress LLaMA-2-7B~\cite{touvron2023llama2} with three base compressors, one per primitive: whitening-based low-rank decomposition (SVD-LLM~\cite{wang2025svdllm}), one-shot activation-aware pruning (Wanda~\cite{sun2024wanda}), and post-training quantization (GPTQ~\cite{frantar2023gptq}, per-layer integer bit-widths in \([2, 8]\)).
The full matrix is replicated on LLaMA-3-8B~\cite{dubey2024llama3}.
A fourth base compressor, FWSVD~\cite{hsulanguage} on BERT\textsubscript{base}~\cite{devlin2019bert}, drives the controlled ablations of Section~\ref{sec:exp:family2} (full tables in Appendices~\ref{sec:appendix-sweep}--\ref{sec:appendix-2}).
All allocations are applied on top of the \emph{identical} base compressor at matched parameter count, so differences are attributable to the allocation alone; ``kept'' denotes the kept parameter fraction.

\vspace{5pt}
\noindent \textbf{Allocation baselines.}
We compare against
(i) \textbf{uniform} allocation, the default of all base compressors;
(ii) \textbf{Bolaco}~\cite{ji2024bolaco}, Bayesian search over per-projection ratios (50 full-model evaluations), rescaled to our budget; and (iii) \textbf{FLRC}~\cite{lu2025flrc}, a gradient-proportional rule; (iv) \textbf{OWL}~\cite{yin2024owl} outlier ratios and (v) MoDeGPT's Block Influence (\textbf{BI})~\cite{lin2025modegpt}; all are transplanted onto the identical base so that only the allocation differs. On Wanda and GPTQ we also report the \emph{bound-driven} knapsack of Theorem~\ref{thm:loss_bound} (Fisher trace \(\times\) measured per-layer distortion; HAWQ-style for bits, BDS for sparsity).

\vspace{5pt}
\noindent \textbf{Metrics and protocol.}
For language models we report WikiText-2~\cite{merity2017pointer} perplexity (PPL\(\downarrow\), sequence length 2048) and the average accuracy over seven zero-shot commonsense-reasoning tasks (ZS\(\uparrow\), lm-evaluation-harness v0.4); ZS cells are 5-seed means where seeded, single-calibration otherwise.
The Fisher estimate requires one backward pass over 128 WikiText-2 calibration sequences of length 2048; the allocation itself takes seconds.
\emph{Allocation cost:} on a single 48\,GB GPU the Fisher pass takes minutes and is shared across all rates and primitives, whereas in our runs Bolaco's Bayesian search requires 50 compress-and-evaluate rounds (\(\sim\)4.5 GPU-hours) and FLRC a full gradient-accumulation pass (\(\sim\)1.5 GPU-hours); at matched parameter count the compressed architectures are identical across allocations, so inference runtime and memory do not depend on the allocation method.
\emph{Seed protocol:} headline comparisons are repeated over five seeds (n = 5; n = 4 where marked) that vary both the Fisher estimate and the compression calibration; we report mean \(\pm\) std and the \emph{paired} per-seed difference \(\Delta\) (DIFW \(-\) uniform), since both allocations share each seed's calibration.

\subsection{Allocator Battle on LLMs}
\label{sec:exp:family1}




\noindent \textbf{Results.}
Table~\ref{tab:svd-battle} is the low-rank battle: eight allocations on the identical SVD-based compressor at matched size.
The closed-form Fisher pair is best in five of six columns --- ahead of Bolaco's 50-round Bayesian search and every heuristic --- at the cost of one backward pass; block granularity wins the aggressive rates, layer the mild ones, and Bolaco takes a single LLaMA-3 cell.
LLaMA-3 is drastically more brittle to low-rank than LLaMA-2 (kept 0.6: 71--73 vs.\ 16 PPL).
Recovery preserves the ranking: after SVD-LLM's closed-form update DIFW\textsubscript{Fair} (block) still leads (48.82 vs.\ 53.20 at kept 0.4; Appendix Table~\ref{tab:svd-recovery}).

Table~\ref{tab:wanda-battle} moves the battle to sparsity, where the picture is regime-dependent: DIFW\textsubscript{Fair} wins or ties the mild cells, but OWL's outlier ratio owns the aggressive kept-0.4 cells (9.19 vs.\ 10.42 on LLaMA-2; 22.70 vs.\ 28.60 on LLaMA-3; outliers, not loss sensitivity, drive damage under heavy removal) while OWL falls \emph{below uniform} at mild LLaMA-3 rates and Bolaco/FLRC trail everywhere (Findings (iv)--(v) analyze DTFMW and BDS).

\begin{table*}[t]
\centering
\caption{\textbf{Allocation battle on GPTQ (quantization)}, per-layer mixed precision, WikiText-2 PPL\(\downarrow\), mean {\tiny$\pm$} std over 5 seeds (\(n=4\) for the LLaMA-2 non-profile rows). Cells: PPL {\tiny$\pm$std} / ZS; best per column in bold. ``\(\equiv\) unif.''\ = provably identical integer bit maps on LLaMA-2 (GQA breaks the degeneracy on LLaMA-3); 2-bit breaks for every allocation and is omitted.}
\label{tab:gptq-battle}
\resizebox{\textwidth}{!}
{
\begin{tabular}{lrrrr}
\toprule
& \multicolumn{2}{c}{\textbf{LLaMA-2}} & \multicolumn{2}{c}{\textbf{LLaMA-3}} \\
\cmidrule(lr){2-3} \cmidrule(lr){4-5}
\textbf{Allocation} & \textbf{3-bit} & \textbf{4-bit} & \textbf{3-bit} & \textbf{4-bit} \\
\midrule
\multicolumn{5}{l}{\emph{Baselines}} \\
uniform & 10.59 {\tiny$\pm$0.72} / 46.0 & 6.10 {\tiny$\pm$0.02} / 53.8 & 401.6 {\tiny$\pm$317.7} / 34.6 & 1497 {\tiny$\pm$797} / 53.8 \\
Bolaco~\cite{ji2024bolaco} & 123.8 {\tiny$\pm$94.5} / 42.2 & 16.23 {\tiny$\pm$9.24} / 54.3 & 352.6 {\tiny$\pm$143.6} / 40.6 & 1508 {\tiny$\pm$784} / 49.7 \\
FLRC~\cite{lu2025flrc} & 42.83 {\tiny$\pm$2.69} / 50.0 & 10.51 {\tiny$\pm$0.12} / 55.9 & 803.4 {\tiny$\pm$65.2} / 30.0 & 15.59 {\tiny$\pm$0.19} / 50.2 \\
OWL~\cite{yin2024owl} & 10.84 {\tiny$\pm$0.79} / 45.7 & 6.32 {\tiny$\pm$0.02} / 54.3 & 551.7 {\tiny$\pm$422.2} / 33.3 & 1540 {\tiny$\pm$895} / 51.9 \\
BI (MoDeGPT)~\cite{lin2025modegpt} & 14.02 {\tiny$\pm$1.85} / 44.1 & 6.18 {\tiny$\pm$0.03} / 53.3 & 2411 {\tiny$\pm$1594} / 33.5 & 25.33 {\tiny$\pm$6.93} / 56.6 \\
\midrule
\multicolumn{5}{l}{\emph{Ours (Fisher-guided)}} \\
DIFW\textsubscript{Fair} (L) & \(\equiv\) unif. & \(\equiv\) unif. & 171.8 {\tiny$\pm$117.8} / 41.9 & 21.06 {\tiny$\pm$8.86} / 56.0 \\
DIFW\textsubscript{Fair} (B) & 10.49 {\tiny$\pm$1.07} / 46.6 & 6.08 {\tiny$\pm$0.03} / 53.7 & 1065 {\tiny$\pm$1576} / 40.3 & 13.02 {\tiny$\pm$8.11} / 56.4 \\
DTFMW (L) & 1585 {\tiny$\pm$1928} / 33.2 & 79.83 {\tiny$\pm$60.60} / 38.8 & 11681 {\tiny$\pm$8430} / 29.6 & 1580 {\tiny$\pm$1421} / 31.0 \\
DTFMW (B) & 306.9 {\tiny$\pm$243.4} / 33.7 & 7.43 {\tiny$\pm$0.50} / 54.9 & 1332 {\tiny$\pm$867} / 32.0 & 250.8 {\tiny$\pm$503.0} / 50.1 \\
HAWQ-style (bound-driven)~\cite{dong2020hawqv2} & \best{8.46 {\tiny$\pm$0.37} / 47.2} & \best{5.94 {\tiny$\pm$0.09} / 53.9} & \best{19.34 {\tiny$\pm$0.38} / 45.4} & \best{9.38 {\tiny$\pm$0.06} / 57.7} \\
\bottomrule
\end{tabular}
}
\end{table*}

Table~\ref{tab:gptq-battle} completes the triad on quantization, where per-layer distortion is most heterogeneous (budget axis: average bits; 3-bit \(\approx\) kept 0.19, 4-bit \(\approx\) 0.25).
The bound-driven knapsack --- our Fisher trace times each layer's \emph{measured} quantization error --- is best in all four columns: on LLaMA-2 it beats every alternative (8.46 / 5.94); on LLaMA-3 it rescues 3-bit outright (19.34 PPL and +10.8 zero-shot points vs.\ uniform's collapse to 402 PPL at near-chance reasoning), and at 4-bit restores near-dense perplexity (9.38 vs.\ 1497, +3.9 points; this collapse is perplexity-dominant: uniform still reasons at 53.8); OWL does not transfer to bits at all.

Paired per-seed statistics and seed-robust zero-shot versions of all headline cells (the LLaMA-3 rescue is worth \(+10.8\) reasoning points at 3-bit) are in Appendices~\ref{sec:appendix:llama3}--\ref{sec:appendix:fullgrid}.

\subsection{Controlled Ablations on BERT}
\label{sec:exp:family2}

\begin{figure}[t]
\centering
\includegraphics[width=\columnwidth]{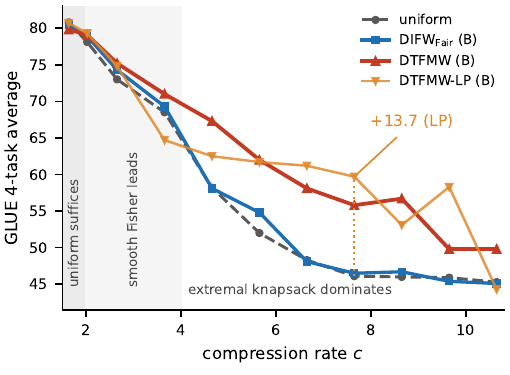}
\caption{\textbf{The three allocation regimes} on the BERT\textsubscript{base} rate sweep (FWSVD base, four-task GLUE average after fine-tuning; full sweep in Table~\ref{tab:rate-sweep}).}
\label{fig:rate-sweep}
\end{figure}

The BERT ablations isolate \emph{where} allocation helps: FWSVD on BERT\textsubscript{base} lets every allocation variant be swept across the full compression range (tables in Appendix~\ref{sec:appendix-sweep}).
The sweep (Figure~\ref{fig:rate-sweep}) exposes three regimes: below \(2\times\) compression uniform is near-lossless; from \(2\times\) to \(3.65\times\) Fisher-guided allocation leads modestly (+1.5 to +2.5; DIFW\textsubscript{Fair} 79.5 vs.\ 78.1 at \(c \approx 2\), block DTFMW from \(c = 2.65\)); and from \(4.65\times\) to \(9.65\times\) the extremal knapsack family (DTFMW and its linear relaxation DTFMW-LP) dominates by +9.0 to +13.7 points, with DTFMW-LP still reaching 54.9 vs.\ 44.1 at an extreme \(\sim\)11.5\(\times\) budget.
Layer-granularity DTFMW is unstable at every rate (anticipating Finding (iv)); a one-shot Wanda-BERT study replicates the block advantage under sparsity (Appendix~\ref{sec:appendix-wanda-bert}).

\subsection{Findings}
\noindent \textbf{(i) A closed-form Fisher rule beats searched and proportional allocation --- and the win is seed-robust where compression bites.}
On the identical base at matched size, DIFW\textsubscript{Fair} (at its better granularity) outperforms Bolaco's search, FLRC, and uniform in all but one column (Table~\ref{tab:svd-battle}; Bolaco takes LLaMA-3 kept 0.6) at the cost of one backward pass; the paired protocol (Table~\ref{tab:llm-seeds}) shows the advantage holds on every seed where compression bites.

\vspace{5pt}
\noindent \textbf{(ii) The value of allocation grows with compression aggressiveness --- and vanishes at saturation.}
On SVD-LLM the advantage over uniform grows from a tie at kept 0.8 to \(-8.9\) PPL one-shot (\(-4.4\) after the update; both seed-robust) at kept 0.4, and the same pattern holds for Wanda and the BERT sweep; at near-lossless rates there is nothing to reallocate (Wanda 0.6/0.8: exact ties, Table~\ref{tab:llm-seeds}).

\vspace{5pt}
\noindent \textbf{(iii) Block granularity is the robust choice where compression bites.}
At aggressive rates block beats layer on the seeded battles: SVD-LLM at kept 0.4 (72.58 vs.\ 78.51) and Wanda at kept 0.4 (10.42 vs.\ 10.51).
On GPTQ, only block granularity can express the allocation on LLaMA-2 (the layer rule is provably identical to uniform under integer bits); GQA breaks this degeneracy on LLaMA-3, where block is the stronger 4-bit rescue (13.02 vs.\ 21.06).
At mild rates layer edges block; one shared ratio per block averages Fisher noise and bounds ratio spread.

\vspace{5pt}
\noindent \textbf{(iv) The extremal knapsack family is the strongest allocator at small scale and high compression, but needs structural regularization at large scale.}
On BERT, the block-granularity knapsack family dominates for \(c \geq 2.65\), peaking at +13.7 points over uniform via the linear relaxation DTFMW-LP (Figure~\ref{fig:rate-sweep}); on billion-parameter models the unregularized extremal solution collapses under every primitive (Tables~\ref{tab:svd-battle}--\ref{tab:gptq-battle}), exactly as the bound predicts: the bang-bang optimum leaves its validity region (Appendix~\ref{sec:theory:assumptions}).
The floor and block granularity of Section~\ref{sec:method:dtfmw} restore stability but not the lead: floored DTFMW (B) swings \(\pm 51\) PPL at LLaMA-2 Wanda 0.4, yet is the seed-robust best where the approximation holds (LLaMA-3 Wanda 0.8: 8.25 \(\pm\) 0.01).

\vspace{5pt}
\noindent \textbf{(v) Allocation signals specialize; the Fisher family is the only one that fails nowhere.}
Each modern baseline wins at most its home primitive (Tables~\ref{tab:wanda-battle}--\ref{tab:gptq-battle}): OWL owns aggressive sparsity but trails uniform in every GPTQ column, while the full second-order objective (Fisher trace \(\times\) measured distortion) is best in all four quantization columns --- our Theorem~\ref{thm:loss_bound} instantiated exactly, with DIFW\textsubscript{Fair} as its distortion-agnostic simplification.
A bound-driven sparsity experiment (BDS) delimits the boundary: the same bound-driven knapsack applied to sparsity \emph{fails} (52.6 vs.\ 10.8 PPL at LLaMA-2 kept 0.4, n = 5), because deep per-layer pruning leaves the quadratic validity region while moderate bit-width perturbations do not.
Practical rule: the knapsack for bits, OWL for aggressive sparsity, DIFW\textsubscript{Fair} otherwise.

\section{Conclusion}
We proposed FI-guided strategies that allocate layer-wise compression ratios from one backward pass and showed they are base-compressor agnostic: at matched size, DIFW\textsubscript{Fair} beats uniform, searched, and proportional allocation across three base compressors, while DTFMW is strongest at small scale and high compression but needs structural regularization at LLM scale.


\section{Limitations}
While the proposed allocation improves on uniform and searched alternatives across three compression primitives on LLM backbones, several limitations remain.

First, using FI as a guiding metric, though effective, may not capture all forms of layer importance; future work could explore alternative signals for compression sensitivity.

Second, the loss bound of Theorem~\ref{thm:loss_bound} is a controlled second-order \emph{estimate} rather than a strict one-sided bound: the Fisher--Hessian equivalence, the block-diagonal structure, and the input-side factorization are approximations without a definite sign --- dropping cross-layer blocks of a positive semi-definite matrix does not by itself produce an upper bound --- and the cubic remainder is absorbed into the inequality.
Appendix~\ref{sec:theory:assumptions} discusses when each step is accurate; empirically, the estimate predicts the relative ordering of allocations only within its validity region (Section~\ref{sec:experiments}).

Third, the closed-form rule is provably optimal only under distortion homogeneity (Proposition~\ref{prop:fisher_sufficient}), and this condition fails at scale.

Figure~\ref{fig:homogeneity} (Section~\ref{sec:theory:general}) measures the condition on our two testbeds, using the Fisher-weighted spectra the allocators actually consume.
On BERT\textsubscript{base} the normalized retained-FI curves form a reasonably tight band (maximum vertical spread 0.70 overall, 0.28 within a projection type), so the homogeneous regime is a fair approximation and the share-based closed form inherits most of the guarantee.
On LLaMA-2-7B homogeneity fails outright (0.93 / 0.66): the curves span nearly the full unit range, and the spread persists even within a single projection type.
In this regime no scalar rule --- share- or density-based --- can reproduce the optimum of the allocation problem; yet trusting the measured spectra exactly is precisely what makes the extremal solution fragile once the quadratic model degrades (Assumption 4, Appendix~\ref{sec:theory:assumptions}).
DIFW\textsubscript{Fair} therefore trades exactness under homogeneity for robustness when it fails, and the experiments show where each side of that trade wins.

Fourth, the proposed intra-layer approach relies on row-wise FI aggregation, which may impose structural constraints that limit flexibility, and because the fair rule's governing equation ignores the original layer's size, it can over-compress a small layer; the rank floor mitigates but does not remove this risk.

Finally, we evaluate on language models up to 8B parameters; extending the allocation across modalities in vision-language models, where gradient scales differ between components, is left for future work.

\bibliography{custom}

\clearpage
\appendix
\label{sec:appendix}
\section{Proofs and Extended Theoretical Analysis}
\label{app:proofs}
This section collects the proofs of the results stated in Section~\ref{sec:theory}, together with the retained-FI form of the bound, its rank-budget optimality, the accuracy extension, and a discussion of assumptions.

\subsection{The Fisher Information Matrix}
\label{app:fim}
Writing $\mathbf{s}(x,y) = \nabla_\theta \log p_{\theta^*}(y \mid x)$ for the per-sample score vector, the Fisher Information Matrix used in Section~\ref{sec:theory:fi} is
\begin{equation}
    \mathbf{F}(\theta^*) = \mathbb{E}_{(x,y) \sim \mathcal{D}} \!\left[ \mathbf{s}(x,y) \, \mathbf{s}(x,y)^\top \right]\!,
    \label{eq:fim_def}
\end{equation}
whose diagonal is estimated on $D$ by the observed FI of Equation~\eqref{eq:observed_fi}.

\subsection{Proof of Theorem~\ref{thm:loss_bound}}
\label{app:proof:thm1}
\begin{proof}
We perform a second-order Taylor expansion of $\mathcal{L}$ around $\theta^*$:
\begin{multline}
    \mathcal{L}(\widetilde{\theta}) = \mathcal{L}(\theta^*) + \underbrace{\nabla_\theta \mathcal{L}(\theta^*)^\top \delta\theta}_{=\,0 \text{ (local minimum)}} \\
    + \frac{1}{2} \delta\theta^\top \, \mathbf{H}(\theta^*) \, \delta\theta + \mathcal{O}\!\left(\|\delta\theta\|^3\right).
    \label{eq:taylor}
\end{multline}
Since $\theta^*$ is a local minimum, the first-order term vanishes. This is a standard assumption in the pruning literature, used in Optimal Brain Damage~\cite{lecun1989optimal} and Optimal Brain Surgeon~\cite{hassibi1993optimal}, and is well-justified for fine-tuned models.
 
Applying the Fisher--Hessian equivalence~\eqref{eq:fisher_hessian}, we obtain:
\begin{equation}
    \Delta\mathcal{L} \approx \frac{1}{2} \, \delta\theta^\top \, \mathbf{F}(\theta^*) \, \delta\theta + \mathcal{O}\!\left(\|\delta\theta\|^3\right).
    \label{eq:loss_fisher}
\end{equation}
 
Under the block-diagonal approximation~\eqref{eq:block_diag}, the quadratic form decomposes across layers:
\begin{equation}
    \delta\theta^\top \, \mathbf{F}(\theta^*) \, \delta\theta = \sum_{\ell=1}^{L} \text{vec}(\delta\mathbf{W}_\ell)^\top \, \mathbf{F}_\ell \, \text{vec}(\delta\mathbf{W}_\ell).
    \label{eq:decompose}
\end{equation}
 
We now reduce each layer's quadratic form to the input side.
Under the Kronecker factorization of Section~\ref{sec:theory:fi}, the layer block is $\mathbf{F}_\ell \approx \mathbf{A}_\ell \otimes \mathbf{B}_\ell$ with $\mathbf{A}_\ell$ the input second-moment factor and $\mathbf{B}_\ell$ the output-gradient factor; using $(\mathbf{A} \otimes \mathbf{B})\,\text{vec}(\mathbf{X}) = \text{vec}(\mathbf{B} \mathbf{X} \mathbf{A}^\top)$ and the symmetry of both factors,
\begin{equation}
    \text{vec}(\delta\mathbf{W}_\ell)^\top (\mathbf{A}_\ell \otimes \mathbf{B}_\ell) \, \text{vec}(\delta\mathbf{W}_\ell) = \left\| \mathbf{B}_\ell^{1/2} \, \delta\mathbf{W}_\ell \, \mathbf{A}_\ell^{1/2} \right\|_F^2.
    \label{eq:kron_identity}
\end{equation}
Taking the identity output factor $\mathbf{B}_\ell \propto \mathbf{I}$ (Assumption 3; FWSVD's row-sum simplification, with the scale absorbed into $\mathbf{A}_\ell$) gives
\begin{equation}
    \text{vec}(\delta\mathbf{W}_\ell)^\top \, \mathbf{F}_\ell \, \text{vec}(\delta\mathbf{W}_\ell) = \left\| \delta\mathbf{W}_\ell \, \mathbf{F}_\ell^{1/2} \right\|_F^2,
    \label{eq:frobenius_identity}
\end{equation}
with $\mathbf{F}_\ell^{1/2} = \mathbf{A}_\ell^{1/2}$ the $n_\ell \times n_\ell$ input-side factor.
We note that without the identity-output-factor step, the correct expression is the two-sided form~\eqref{eq:kron_identity}; the one-sided form is exactly the simplification that makes the input-side spectra of Definition~\ref{def:fi_ratio} computable.
Combining~\eqref{eq:loss_fisher},~\eqref{eq:decompose}, and~\eqref{eq:frobenius_identity} yields the result.
\end{proof}

\subsection{Proof of Proposition~\ref{prop:general_alloc}}
\label{app:proof:gen}
\begin{proof}
Each $D_\ell$ extends continuously to $r = 0$ with the finite value $D_\ell(0) = \mathrm{FI}_\ell$, so Problem~\eqref{eq:general_alloc} minimizes a separable convex objective over the compact convex set $\{r \in [0,1]^L : \sum_\ell s_\ell r_\ell \leq B\}$ (or $[r_{\min}, 1]^L$ when a floor is imposed); a minimizer therefore exists, and since all constraints are affine, the KKT conditions are both necessary and sufficient.
Writing the Lagrangian $\sum_\ell D_\ell(r_\ell) + \lambda\left(\sum_\ell s_\ell r_\ell - B\right)$ with multiplier $\lambda \geq 0$, stationarity in $r_\ell$ gives $D_\ell'(r_\ell^*) + \lambda s_\ell = 0$ whenever $r_\ell^*$ lies in the interior of $(0,1]$, that is, $-D_\ell'(r_\ell^*) / s_\ell = \lambda$.
Since each $D_\ell$ is non-increasing, $-D_\ell' \geq 0$ and hence $\lambda \geq 0$; the budget constraint is tight whenever the budget actually forces compression ($B < \sum_\ell s_\ell$) and at least one $D_\ell$ is strictly decreasing.
\end{proof}

\subsection{Proof of Proposition~\ref{prop:fisher_sufficient}}
\label{app:proof:suff}
\begin{proof}
Substituting $D_\ell = \mathrm{FI}_\ell \cdot d$ into the stationarity condition of Proposition~\ref{prop:general_alloc} yields $\mathrm{FI}_\ell \, d'(r_\ell^*) + \lambda s_\ell = 0$, hence
\begin{equation}
    d'(r_\ell^*) = -\lambda \, \frac{s_\ell}{\mathrm{FI}_\ell} .
    \label{eq:suff_stationarity}
\end{equation}
Because $d$ is strictly convex, $d'$ is strictly increasing and therefore invertible, so $r_\ell^* = (d')^{-1}\!\left(-\lambda \, s_\ell / \mathrm{FI}_\ell\right)$.
Writing $u_\ell = \mathrm{FI}_\ell / s_\ell$ and $\varphi(u) = (d')^{-1}(-\lambda / u)$ gives~\eqref{eq:fisher_rule}.
The map $u \mapsto -\lambda / u$ is increasing for $\lambda > 0$ and $(d')^{-1}$ is increasing, so $\varphi$ is increasing.
The multiplier $\lambda$ is determined by the budget $\sum_\ell s_\ell r_\ell^* = B$ and is shared by all layers, so $\varphi$ does not depend on $\ell$.
\end{proof}

\subsection{Low-Rank Instantiation of the Bound}
\label{app:proof:lem1}
For low-rank truncation, the Fisher-weighted norm of Theorem~\ref{thm:loss_bound} reduces to the discarded Fisher-weighted singular values. Here Fisher-weighted SVD refers to the operation $\mathrm{SVD}(\mathbf{W}_\ell^* \mathbf{F}_\ell^{1/2})$ that DTFMW ranks directions by, rather than to any particular compressor.
Throughout we assume $\mathbf{F}_\ell \succ 0$, so that $\mathbf{F}_\ell^{-1/2}$ exists and the truncation is recoverable in the original coordinates; in practice, zero Fisher entries are clamped to a small floor.

\begin{lemma}[Fisher-Weighted Reconstruction Error]
\label{lem:fwsvd_error}
Let $\mathbf{W}_\ell^*$ have the Fisher-weighted SVD $\mathbf{W}_\ell^* \mathbf{F}_\ell^{1/2} = \sum_{i=1}^{R_\ell} \widetilde{\sigma}_{i,\ell} \, \mathbf{u}_{i,\ell} \, \mathbf{v}_{i,\ell}^\top$, where $\widetilde{\sigma}_{1,\ell} \geq \widetilde{\sigma}_{2,\ell} \geq \cdots \geq \widetilde{\sigma}_{R_\ell,\ell} \geq 0$ are the Fisher-weighted singular values and $R_\ell = \min(m_\ell, n_\ell)$. If $\widetilde{\mathbf{W}}_\ell$ is the best rank-$k_\ell$ approximation under the Fisher-weighted norm, then:
\begin{equation}
    \left\| \delta\mathbf{W}_\ell \, \mathbf{F}_\ell^{1/2} \right\|_F^2 = \sum_{i=k_\ell+1}^{R_\ell} \widetilde{\sigma}_{i,\ell}^2.
    \label{eq:truncation_error}
\end{equation}
\end{lemma}

\begin{proof}
By the Eckart--Young--Mirsky theorem~\cite{eckart1936approximation}, the optimal rank-$k_\ell$ approximation of $\mathbf{W}_\ell^* \mathbf{F}_\ell^{1/2}$ in the Frobenius norm is obtained by retaining the top $k_\ell$ singular value components. The corresponding approximation $\widetilde{\mathbf{W}}_\ell \mathbf{F}_\ell^{1/2} = \sum_{i=1}^{k_\ell} \widetilde{\sigma}_{i,\ell} \, \mathbf{u}_{i,\ell} \, \mathbf{v}_{i,\ell}^\top$ yields a residual whose squared Frobenius norm equals the sum of the squared discarded singular values.
\end{proof}

Substituting Lemma~\ref{lem:fwsvd_error} into Theorem~\ref{thm:loss_bound}, we obtain:

\begin{corollary}[Loss Bound in Terms of Discarded Fisher-Weighted Singular Values]
\label{cor:loss_sv}
Under the assumptions of Theorem~\ref{thm:loss_bound}, the increase in task loss satisfies:
\begin{equation}
    \Delta\mathcal{L} \leq \frac{1}{2} \sum_{\ell=1}^{L} \sum_{i=k_\ell+1}^{R_\ell} \widetilde{\sigma}_{i,\ell}^2 + \mathcal{O}\!\left(\|\delta\theta\|^3\right).
    \label{eq:loss_bound_sv}
\end{equation}
\end{corollary}

\label{sec:theory:retained_fi}
 
For low-rank truncation the bound becomes concrete. Writing $\widetilde{\sigma}_{i,\ell}$ for the singular values of $\mathbf{W}_\ell^* \mathbf{F}_\ell^{1/2}$, the Fisher-weighted spectrum that DTFMW ranks directions by, the per-layer norm in Theorem~\ref{thm:loss_bound} equals the sum of the discarded $\widetilde{\sigma}_{i,\ell}^2$ (Lemma~\ref{lem:fwsvd_error} and Corollary~\ref{cor:loss_sv}, Appendix~\ref{app:proof:lem1}).
We now express that bound in terms of the fraction of retained Fisher Information.
 
\begin{definition}[Layer-wise Fisher Information and Retention Ratio]
\label{def:fi_ratio}
Let $R_\ell = \min(m_\ell, n_\ell)$ denote the full rank of layer $\ell$ (we reserve lowercase $r$ for kept fractions).
The total Fisher Information at layer $\ell$ is defined as:
\begin{equation}
    \mathrm{FI}_\ell \triangleq \sum_{i=1}^{R_\ell} \widetilde{\sigma}_{i,\ell}^2 = \left\| \mathbf{W}_\ell^* \, \mathbf{F}_\ell^{1/2} \right\|_F^2.
    \label{eq:fi_total}
\end{equation}
The retained Fisher Information after rank-$k_\ell$ compression is $\mathrm{FI}_\ell^{\mathrm{ret}} = \sum_{i=1}^{k_\ell} \widetilde{\sigma}_{i,\ell}^2$, and the \textit{FI retention ratio} is:
\begin{equation}
    \rho_\ell \triangleq \frac{\mathrm{FI}_\ell^{\mathrm{ret}}}{\mathrm{FI}_\ell} = \frac{\sum_{i=1}^{k_\ell} \widetilde{\sigma}_{i,\ell}^2}{\sum_{i=1}^{R_\ell} \widetilde{\sigma}_{i,\ell}^2} \in [0, 1].
    \label{eq:retention_ratio}
\end{equation}
\end{definition}
 
Using Definition~\ref{def:fi_ratio}, the discarded FI at layer $\ell$ equals $(1 - \rho_\ell) \cdot \mathrm{FI}_\ell$. Substituting into the low-rank form of the bound (Corollary~\ref{cor:loss_sv}):
 
\begin{corollary}[Loss Bound via FI Retention Ratio]
\label{cor:loss_retention}
\begin{equation}
    \Delta\mathcal{L} \leq \frac{1}{2} \sum_{\ell=1}^{L} (1 - \rho_\ell) \cdot \mathrm{FI}_\ell + \mathcal{O}\!\left(\|\delta\theta\|^3\right).
    \label{eq:loss_bound_retention}
\end{equation}
\end{corollary}
 
\noindent
\textbf{Interpretation.}
Equation~\eqref{eq:loss_bound_retention} reveals that the task loss degradation is controlled by a \emph{weighted sum} of the discarded FI fractions across layers, where each layer's contribution is weighted by its total FI magnitude $\mathrm{FI}_\ell$. This provides a direct theoretical justification for our proposed layer-wise allocation strategies:
\begin{itemize}[nosep]
    \item Layers with larger $\mathrm{FI}_\ell$ contribute more heavily to the bound and should receive higher ranks (less compression), as even a small fraction of discarded FI results in a large absolute contribution.
    \item For a fixed global rank budget $K = \sum_{\ell=1}^{L} k_\ell$, the optimal allocation minimizes $\sum_\ell (1 - \rho_\ell) \cdot \mathrm{FI}_\ell$, which is equivalent to maximizing $\sum_\ell \rho_\ell \cdot \mathrm{FI}_\ell$ (the total retained FI).
\end{itemize}

\subsection{Optimality of FI-Guided Rank Allocation}
\label{app:proof:prop1}
\label{sec:theory:optimality}
 
We now formalize the optimal rank allocation problem and connect it to our proposed methods.
 
\begin{proposition}[Optimal Rank Allocation under Global Budget]
\label{prop:optimal_alloc}
Given a global rank budget $K = \sum_{\ell=1}^{L} k_\ell$ and layer-wise FI spectra $\{\widetilde{\sigma}_{i,\ell}\}_{i,\ell}$, the rank allocation $\{k_\ell^*\}_{\ell=1}^{L}$ that minimizes the loss bound~\eqref{eq:loss_bound_sv} is:
\begin{equation}
    \{k_1^*, \ldots, k_L^*\} = \underset{\substack{k_1, \ldots, k_L \geq 0 \\ \sum_\ell k_\ell = K}}{\arg\min} \; \sum_{\ell=1}^{L} \sum_{i=k_\ell+1}^{R_\ell} \widetilde{\sigma}_{i,\ell}^2.
    \label{eq:optimal_alloc}
\end{equation}
This is equivalent to selecting the top-$K$ Fisher-weighted singular values across all layers, which corresponds to a global greedy allocation.
\end{proposition}

\noindent
\textbf{Remark.} The two methods proposed in Section~\ref{sec:Method} sit at opposite ends of the design space that~\eqref{eq:optimal_alloc} defines.
\textit{DTFMW} solves~\eqref{eq:optimal_alloc} directly via the same greedy selection, with the budget measured in parameters.
\textit{DIFW\textsubscript{Fair}} is \emph{not} an approximation of this extremal optimum but a smooth surrogate motivated by the monotonicity of Corollary~\ref{cor:loss_retention}: it moves every layer's rate a bounded distance from uniform, in the direction the bound prescribes.
Section~\ref{sec:experiments} shows this distinction is not cosmetic --- the extremal optimum is the stronger allocator exactly while the quadratic model remains valid (small models, moderate rates), and the smooth surrogate is the robust one when it does not (billion-parameter models, aggressive rates).

\noindent \textbf{Proof.}
\begin{proof}
The objective in~\eqref{eq:optimal_alloc} can be rewritten as:
\begin{equation}
    \sum_{\ell=1}^{L} \sum_{i=k_\ell+1}^{R_\ell} \widetilde{\sigma}_{i,\ell}^2 = \underbrace{\sum_{\ell=1}^{L} \mathrm{FI}_\ell}_{\text{constant}} \;-\; \sum_{\ell=1}^{L} \sum_{i=1}^{k_\ell} \widetilde{\sigma}_{i,\ell}^2.
    \label{eq:rewrite}
\end{equation}
Since $\sum_\ell \mathrm{FI}_\ell$ is constant with respect to $\{k_\ell\}$, minimizing~\eqref{eq:rewrite} is equivalent to maximizing $\sum_{\ell} \sum_{i=1}^{k_\ell} \widetilde{\sigma}_{i,\ell}^2$. 
Because the Fisher-weighted singular values within each layer are sorted in decreasing order, this is a fractional knapsack problem~\cite{frantar2022spdy}: we assign rank by iteratively selecting the next-largest $\widetilde{\sigma}_{i,\ell}^2$ across all layers until the budget $K$ is exhausted. The greedy solution is provably optimal for fractional knapsack~\cite{korte2012combinatorial}.
\end{proof}

\subsection{Accuracy Degradation Bound via KL Divergence}
\label{sec:theory:kl}
 
We now extend the loss degradation bound to a bound on classification accuracy using the information-theoretic relationship between FI and KL divergence.
 
\begin{theorem}[FI--KL--Accuracy Degradation Chain]
\label{thm:accuracy_bound}
Let $p_{\theta^*}(y \mid x)$ denote the output distribution of the original model and $p_{\widetilde{\theta}}(y \mid x)$ the output distribution of the compressed model.
Under the second-order local approximation and the block-diagonal Fisher assumption, the expected classification error degradation satisfies:
\begin{multline}
    \mathbb{E}_x\!\left[|P_{\mathrm{err}}(\widetilde{\theta}; x) - P_{\mathrm{err}}(\theta^*; x)|\right] \\
    \leq \sqrt{\frac{1}{4} \sum_{\ell=1}^{L} (1 - \rho_\ell) \cdot \mathrm{FI}_\ell} \;+\; \mathcal{O}\!\left(\|\delta\theta\|^{3/2}\right),
    \label{eq:accuracy_bound}
\end{multline}
where $P_{\mathrm{err}}(\theta; x) = 1 - \max_y p_\theta(y \mid x)$ is the per-sample misclassification probability.
\end{theorem}
 
\noindent \textit{Proof in Appendix~\ref{app:proof:thm2}.}
 
\noindent
\textbf{Interpretation.}
Theorem~\ref{thm:accuracy_bound} establishes that \emph{maximizing total retained FI minimizes an upper bound on classification accuracy degradation}. 
This provides a formal justification for both proposed allocation strategies:
any allocation $\{k_\ell\}$ that increases $\sum_\ell \rho_\ell \cdot \mathrm{FI}_\ell$ provably tightens the accuracy bound.
Moreover, the bound reveals a \emph{diminishing returns} structure: because of the square root, doubling the discarded FI increases the accuracy bound by only a factor of $\sqrt{2}$, suggesting robustness to moderate compression.
Two caveats scope the result.
First, $P_{\mathrm{err}}$ is the model's own confidence-based error; it tracks the true task error under the well-specification of Assumption 2 (and then with an additional factor of two via the standard argmax-switching argument).
Second, the chain inherits the second-order caveats of Theorem~\ref{thm:loss_bound}, including the distinction between the model Fisher used in the KL expansion and the observed FI our allocators consume (Appendix~\ref{app:proof:thm2}).

\noindent \textbf{Proof of Theorem~\ref{thm:accuracy_bound}.}
\label{app:proof:thm2}
\begin{proof}
The proof proceeds in three steps.
 
\textbf{Step 1: Local FI--KL equivalence.}
By a second-order Taylor expansion of the KL divergence around $\theta^*$~\cite{kullback1951information, cover2006elements}, for a small perturbation $\delta\theta = \widetilde{\theta} - \theta^*$:
\begin{equation}
    D_{\mathrm{KL}}\!\left(p_{\theta^*}(\cdot \mid x) \,\|\, p_{\widetilde{\theta}}(\cdot \mid x)\right) \approx \frac{1}{2} \, \delta\theta^\top \, \mathbf{F}_x(\theta^*) \, \delta\theta,
    \label{eq:kl_fisher}
\end{equation}
where $\mathbf{F}_x(\theta^*) = \mathbb{E}_{y \sim p_{\theta^*}(\cdot \mid x)}\!\left[\nabla_\theta \log p_{\theta^*}(y \mid x) \, \nabla_\theta \log p_{\theta^*}(y \mid x)^\top\right]$ is the per-sample Fisher matrix.
Note that this is the \emph{model} Fisher, with labels drawn from $p_{\theta^*}$; it coincides with the observed FI of Equation~\eqref{eq:observed_fi} that our allocators consume only under the well-specification of Assumption 2~\cite{kunstner2019limitations}.

Taking expectation over $x$ and applying the block-diagonal and input-side approximations of Section~\ref{sec:theory:fi}:
\begin{multline}
    \mathbb{E}_x\!\left[D_{\mathrm{KL}}\!\left(p_{\theta^*} \,\|\, p_{\widetilde{\theta}}\right)\right] \approx \frac{1}{2} \, \delta\theta^\top \, \mathbf{F}(\theta^*) \, \delta\theta \\
    = \frac{1}{2} \sum_{\ell=1}^{L} (1 - \rho_\ell) \cdot \mathrm{FI}_\ell,
    \label{eq:expected_kl}
\end{multline}
where the last equality follows from Corollary~\ref{cor:loss_retention}.
 
\textbf{Step 2: Pinsker's inequality.}
By Pinsker's inequality~\cite{tsybakov2009introduction}, the total variation distance between two distributions is bounded by the KL divergence:
\begin{multline}
    \mathrm{TV}\!\left(p_{\theta^*}(\cdot \mid x), \, p_{\widetilde{\theta}}(\cdot \mid x)\right) \\
    \leq \sqrt{\frac{1}{2} \, D_{\mathrm{KL}}\!\left(p_{\theta^*}(\cdot \mid x) \,\|\, p_{\widetilde{\theta}}(\cdot \mid x)\right)}.
    \label{eq:pinsker}
\end{multline}
 
\textbf{Step 3: TV--Error coupling.}
For any two distributions $P, Q$ over a discrete label space, the difference in misclassification probabilities under the argmax decision rule is bounded by the total variation distance~\cite{devroye1996probabilistic}:
\begin{equation}
    |P_{\mathrm{err}}(Q) - P_{\mathrm{err}}(P)| \leq \mathrm{TV}(P, Q).
    \label{eq:tv_error}
\end{equation}
Combining~\eqref{eq:expected_kl},~\eqref{eq:pinsker}, and~\eqref{eq:tv_error}, and applying Jensen's inequality:
\begin{align}
    &\mathbb{E}_x\!\left[|P_{\mathrm{err}}(\widetilde{\theta}; x) - P_{\mathrm{err}}(\theta^*; x)|\right] \nonumber \\
    &\quad\leq \mathbb{E}_x\!\left[\sqrt{\frac{1}{2} \, D_{\mathrm{KL}}\!\left(p_{\theta^*} \,\|\, p_{\widetilde{\theta}}\right)}\right] \nonumber \\
    &\quad\leq \sqrt{\frac{1}{2} \, \mathbb{E}_x\!\left[D_{\mathrm{KL}}\!\left(p_{\theta^*} \,\|\, p_{\widetilde{\theta}}\right)\right]} \label{eq:jensen} \\
    &\quad\leq \sqrt{\frac{1}{4} \sum_{\ell=1}^{L} (1 - \rho_\ell) \cdot \mathrm{FI}_\ell}, \nonumber
\end{align}
where~\eqref{eq:jensen} uses Jensen's inequality ($\sqrt{\cdot}$ is concave).
\end{proof}

\subsection{Discussion of Assumptions}
\label{sec:theory:assumptions}

\begin{algorithm}[t]
\caption{DTFMW: greedy Fisher knapsack (layer level)}
\label{alg:dtfmw}
\begin{algorithmic}[1]
\REQUIRE spectra \( \{\sigma_{W,i}^2\}_{W \in Q} \), budget \( B = S_{\text{total}}/c \), floor \( r_{\min} \)
\STATE \( r_W \gets r_{\min} \) for all \( W \); \quad \( s \gets \sum_{W} r_{\min} (m_W + n_W) \)
\STATE \( H \gets \) max-heap of \( \left( \sigma_{W, r_W + 1}^2 / (m_W + n_W),\, W \right) \) over \( Q \)
\WHILE{\( H \) is not empty}
    \STATE pop the densest increment \( (\cdot, W) \) from \( H \)
    \IF{\( s + (m_W + n_W) \leq B \)}
        \STATE \( r_W \gets r_W + 1 \); \quad \( s \gets s + (m_W + n_W) \)
        \IF{\( r_W < \min(m_W, n_W) \)}
            \STATE push \( \left( \sigma_{W, r_W + 1}^2 / (m_W + n_W),\, W \right) \) onto \( H \)
        \ENDIF
    \ENDIF
\ENDWHILE
\RETURN ranks \( \{r_W\} \), i.e., \( c_W = \frac{m_W n_W}{r_W (m_W + n_W)} \)
\end{algorithmic}
\end{algorithm}
 
We summarize the key assumptions underlying our theoretical results and discuss their validity.
 
\begin{enumerate}
    \item \textbf{Local minimum assumption.} We assume that the model is fine-tuned to a local minimum of $\mathcal{L}$, so the gradient vanishes, and that $\mathcal{L}$ has bounded third derivatives in a neighborhood of $\theta^*$, so the cubic remainder in the Taylor expansion is controlled. This is standard in the OBD/OBS pruning literature~\cite{lecun1989optimal, hassibi1993optimal} and is well-justified for fine-tuned BERT models.
    
    \item \textbf{Fisher--Hessian equivalence.} This holds exactly when the model is well-specified, and approximately when training loss is small~\cite{martens2020new, kunstner2019limitations}. For well-trained BERT models on GLUE tasks, this is a reasonable assumption. We note that WoodFisher~\cite{singh2020woodfisher} and the Optimal BERT Surgeon~\cite{kurtic2022optimal} successfully apply this equivalence in practice.
    
    \item \textbf{Block-diagonal and Kronecker Fisher structure.} Following KFAC~\cite{martens2015optimizing, grosse2016kronecker}, we assume cross-layer interactions are negligible. Within a layer we additionally use the Kronecker factorization with an identity output factor (Section~\ref{sec:theory:fi}) --- FWSVD's row-sum simplification --- which is what makes the input-side spectra of Definition~\ref{def:fi_ratio} well-defined; the resulting importance estimates are accurate in practice~\cite{hsulanguage}. We note that EvoPress~\cite{sieberling2024evopress} observed violations of error monotonicity (the assumption that total model error is the sum of layer-wise errors) in LLMs at high compression ratios. However, for BERT-scale models at moderate compression levels, the block-diagonal assumption and error monotonicity are well-supported empirically~\cite{frantar2022spdy}.
    
    \item \textbf{Second-order sufficiency.} The Taylor expansion truncation neglects third- and higher-order terms. This is valid when the perturbation $\delta\theta$ is small, i.e., when the compression is moderate. For aggressive compression, higher-order effects may become significant; however, the second-order bound remains a useful approximation and correctly predicts the relative ordering of allocation strategies.
\end{enumerate}

\subsection{Greedy Implementation of DTFMW}
\label{app:alg}
Algorithm~\ref{alg:dtfmw} implements the greedy knapsack of Section~\ref{sec:method:dtfmw} with a max-heap over per-layer rank increments, ranked by Fisher value per parameter; with \( N = \sum_W \min(m_W, n_W) \) candidate increments it runs in \( \mathcal{O}(N \log N) \) time.

\FloatBarrier
\section{Alternative Allocation Variants}
\label{sec:appendix-variants}
Besides DIFW\textsubscript{Fair} and DTFMW (Section~\ref{sec:Method}), we developed and evaluated four alternative allocation variants.
They are competitive in specific regimes but are dominated overall by the two main methods; we summarize them here and report a controlled comparison in Table~\ref{tab:dtfmw-variants}.

\subsection{DIFW\textsubscript{Total}}
DIFW\textsubscript{Total} is the size-weighted counterpart of DIFW\textsubscript{Fair}: instead of matching the \emph{average} compression ratio, it matches the \emph{total} parameter count exactly, accounting for the dimensions \( m_W \times n_W \) of each weight matrix.
It minimizes the loss:
\begin{equation}
\label{eq:overallinterlayer}
    \mathcal{L}_{\text{Total}}(\alpha) = \left( \frac{S_{\text{Total}}}{c} - \sum_{W \in Q} \frac{m_W \times n_W}{\alpha (1 - p_W)} \right)^2,
\end{equation}
where \( \alpha \) is a shared scalar and \( S_{\text{Total}} \) is the total number of parameters subject to compression, with the analytical solution:
\begin{equation}
    \alpha = \frac{c}{S_{\text{Total}}} \sum_{W \in Q} \frac{m_W \times n_W}{1 - p_W}.
\end{equation}
Per-layer ranks follow from Equation~\ref{eq:calculate-rank-k-from-cw}.
DIFW\textsubscript{Total} performs on par with DIFW\textsubscript{Fair} in our BERT ablations (e.g., a four-task average of 78.8 vs.\ 79.5 at \( c \approx 2 \)) and never clearly above it, so we report DIFW\textsubscript{Fair} in the main paper.

\subsection{FPIFW}
Fixed Percentage Intra-layer Fisher-weighted Compression (FPIFW) retains, within each layer, the smallest number of Fisher-weighted spectral components whose cumulative FI reaches a fixed retention threshold, so the per-layer compression ratio is inferred rather than assigned.
Concretely, the row-summed FI vector \( \tilde{I}_{W_{\text{Reduced}}} \), with \( \tilde{I}_{W_{\text{Reduced}_i}} = \sum_{j} \widetilde{\mathcal{I}_\theta}_{W_{i,j}} \), is projected onto the right singular-vector space as \( \tilde{I}_{W_\text{Projected}} = V^\top \, \text{DIAG}(\tilde{I}_{W_\text{Reduced}}) \, V \); the singular values are reweighted element-wise, \( \Sigma_{\text{Weighted}} = \Sigma \odot \tilde{I}_{W_\text{Projected}} \); and the top-\( k \) components whose cumulative sum meets the threshold are retained.
Because FPIFW offers no direct control over the resulting global compression ratio, size-matched comparisons require model-specific tuning of the threshold, and its GLUE results trail the inter-layer methods (Table~\ref{Tab:GLUE-BERT-66M}).

\subsection{DTFMW-LP: The Linear Relaxation of DTFMW}
\label{sec:appendix-dtfmw-lp}
The precursor of the DTFMW knapsack replaces the true retained-FI value \( V_W(r_W) \) of Equation~\ref{eq:dtfmw-value} with the linear proxy \( \widetilde{\mathcal{I}}_W / c_W \), i.e., it assumes each layer's FI is retained in proportion to its kept fraction:
\begin{align}
\min_{C} \; & \mathcal{L}(C) = \widetilde{\mathcal{I}}_{\text{Total}} - \sum_{W \in Q} \frac{\widetilde{\mathcal{I}}_W}{c_W} \label{eq:dtfmw-objective-func} \\
\text{s.t.} \; & c_W \geq 1 \quad \forall\, W \in Q, \nonumber \\
& \sum_{W \in Q} \frac{m_W n_W}{c_W} = \frac{S_{\text{total}}}{c}. \nonumber
\end{align}
In the kept-fraction variable \( y_W = 1/c_W \), both the objective and the budget constraint are linear, so DTFMW-LP is a fractional knapsack solved exactly by spending the budget on layers in decreasing order of the Fisher-per-parameter density \( \widetilde{\mathcal{I}}_W / (m_W n_W) \); it is precisely the DTFMW objective of Equation~\ref{eq:dtfmw-knapsack} with \( V_W \) linearized.
(An interior-point solver~\cite{byrd1999interior} reaches the same optimum but is far slower.)
Because a linear program on a box attains its optimum at a vertex, DTFMW-LP assigns almost every layer either its maximum or its minimum kept fraction; this bang-bang behavior is even more pronounced than in DTFMW, whose decaying spectra yield interior solutions.

\subsection{DTFMW-MC}
DTFMW-MC solves the DTFMW-LP objective by Monte Carlo search instead of exactly: it samples candidate configurations \( C \), enforces the budget through rescaling with a tolerance of \( c \pm 0.3\% \), and keeps the best-scoring feasible sample (100{,}000 samples in our experiments).
Appendix~\ref{sec:appendix-1} visualizes the sampled solution space.
DTFMW-MC approaches, but does not reach, the exact LP optimum, and adds considerable cost.

\subsection{DTFMW-pos}
DTFMW-pos augments DTFMW-LP with a depth prior \( g(d) = \exp\!\left(-\beta \, d / (L-1)\right) \) with \( \beta = 1 \), which favors retaining earlier blocks.
The prior rescues layer granularity (81.9 vs.\ 80.6 without it, Table~\ref{tab:dtfmw-variants}) but is redundant at block granularity, where the block structure already regularizes the allocation.

\begin{table}[t]
\centering
\small
\caption{\textbf{Allocation-variant comparison} on BERT\textsubscript{base} with the FWSVD base at \( c \approx 2.02 \); average primary metric over seven GLUE tasks, after fine-tuning. DTFMW denotes the knapsack of Section~\ref{sec:method:dtfmw}; DTFMW-LP its linear relaxation. DTFMW (layer) fails (CoLA drops to 0) without block structure --- the same fragility observed at the 7B scale.}
\label{tab:dtfmw-variants}
\fitbox
{
\begin{tabular}{lr}
\toprule
\textbf{Variant} & \textbf{7-task Avg} \\
\midrule
DIFW\textsubscript{Fair} & \best{82.7} \\
DTFMW (block) & 82.4 \\
DTFMW-LP (block) & 82.2 \\
DTFMW-pos (block) & 82.0 \\
DTFMW-pos (layer) & 81.9 \\
uniform & 81.9 \\
DTFMW-LP (layer) & 80.6 \\
DTFMW (layer) & 71.4 \\
\bottomrule
\end{tabular}
}
\end{table}

\FloatBarrier
\section{Monte Carlo Sampling of Compression Configurations}
\label{sec:appendix-1}
The feasible space of compression configurations for a given target compression ratio \( c \) is defined over \(|Q|\) dimensions, where each layer \( W \in Q \) is assigned its own compression ratio \( c_W \).
To better understand this space, we simulate a compression scenario using Monte Carlo sampling over a synthetic set of 12 layers \( \tilde{W} \in \tilde{Q} \). 
Each simulated layer \( \tilde{W} \) has randomly sampled dimensions \( m_W, n_W \in [1, 100] \), drawn uniformly. 
Additionally, each layer is assigned a simulated observed FI value \( \tilde{\tilde{I}} \in [0.1, 10.0] \), also sampled from a uniform distribution.
For each simulation, a compression configuration \( \tilde{C} = \{\tilde{c}_W\} \), with \( \tilde{c}_W \in [1, 10] \), is sampled uniformly. A sample is considered acceptable if it results in a total compression ratio \( \tilde{c} \) within \( \pm5\% \) of the target \( c \), where the total compression is defined as:

\begin{equation}
    \tilde{c} =  \sum_{W \in Q} \frac{\tilde{m}_W \times \tilde{n}_W}{\tilde{c}_W \cdot \tilde{S}_{\text{total}}}, \quad \tilde{c}_W \in \tilde{C},\ \tilde{W} \in \tilde{Q}.
\end{equation}

Since uniformly sampling \( \tilde{C} \) often fails to yield configurations satisfying the constraint \( \tilde{c} \approx c \), we introduce a correction mechanism. After each sample, if \( \tilde{c} \neq c \), we rescale all \( \tilde{c}_W \in \tilde{C} \) by a factor \( z = \frac{c}{\tilde{c}} \). If \( \tilde{c} > c \), the factor \( z < 1 \) increases the compression (reduces ranks), and if \( \tilde{c} < c \), then \( z > 1 \) relaxes compression (increases ranks).

\begin{figure}[!htbp]
    \centering
    \includegraphics[width=1.0\linewidth]{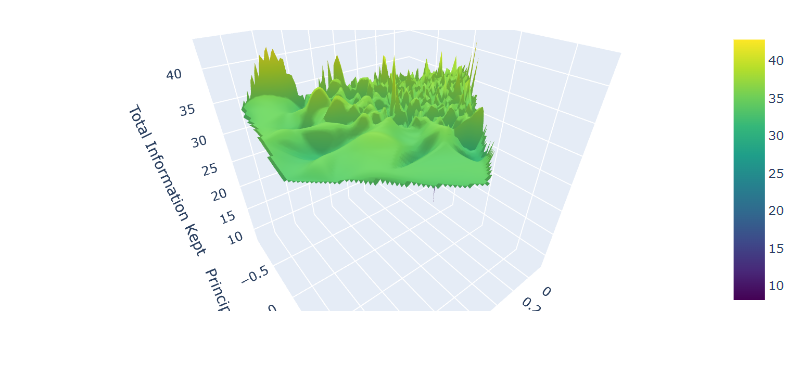}
    \caption{Principal Component Analysis projection of Monte Carlo sampled compression configurations, visualizing compression ratios versus total information retention.}
    \label{fig:plotproblem}
\end{figure}

This sampling procedure generates a large collection of feasible compression configurations. 
To visualize the structure of the solution space, we apply Principal Component Analysis (PCA)~\cite{wold1987principal} to reduce the sampled high-dimensional vectors to two dimensions. 
Figure~\ref{fig:plotproblem} illustrates the complexity of the compression landscape. 
The presence of numerous local optima makes it difficult for standard optimizers such as Adam~\cite{kingma2014adam} or RMSProp~\cite{tieleman2012lecture} to reliably find globally optimal solutions. 
This motivates our use of sampling-based strategies to both explore the space and guide the search toward solutions that preserve maximal information under a fixed compression ratio.

\FloatBarrier
\section{Compression-Rate Sweep on BERT}
\label{sec:appendix-sweep}

\begin{table*}[t]
\centering
\small
\caption{\textbf{Compression-rate sweep} on BERT\textsubscript{base} with the FWSVD base; average primary metric over MRPC/STS-B/CoLA/SST-2 after fine-tuning. L/B = layer/block granularity. DTFMW = knapsack (Section~\ref{sec:method:dtfmw}); DTFMW-LP = linear relaxation (Appendix~\ref{sec:appendix-dtfmw-lp}). The last column reports the margin of the best allocation over uniform.}
\label{tab:rate-sweep}
\fitbox
{
\begin{tabular}{rrrrrrrrr}
\toprule
\( c \) & \textbf{Uniform} & \textbf{DIFW\textsubscript{Fair} (L)} & \textbf{DIFW\textsubscript{Fair} (B)} & \textbf{DTFMW-LP (L)} & \textbf{DTFMW-LP (B)} & \textbf{DTFMW (L)} & \textbf{DTFMW (B)} & \textbf{best\,\(-\)\,unif.} \\
\midrule
1.63 & \best{80.8} & 80.7 & 80.4 & 80.0 & 80.7 & 67.3 & 79.8 & +0.0 \\
2.02 & 78.1 & \best{79.5} & 78.9 & 77.5 & 79.3 & 63.5 & 79.1 & +1.5 \\
2.65 & 73.0 & 73.3 & 74.3 & 61.9 & 74.8 & 61.9 & \best{75.2} & +2.1 \\
3.65 & 68.5 & 69.2 & 69.3 & 43.8 & 64.7 & 44.9 & \best{71.0} & +2.5 \\
4.65 & 58.2 & 52.7 & 58.1 & 42.4 & 62.5 & 43.0 & \best{67.3} & +9.0 \\
5.65 & 52.0 & 45.4 & 54.8 & 44.6 & 61.7 & 51.7 & \best{62.0} & +10.0 \\
6.65 & 48.3 & 46.3 & 48.1 & 43.0 & \best{61.2} & 45.6 & 58.1 & +12.9 \\
7.65 & 46.1 & 48.4 & 46.5 & 43.4 & \best{59.7} & 45.5 & 55.8 & +13.7 \\
8.65 & 46.0 & 44.9 & 46.7 & 43.5 & 53.1 & 44.0 & \best{56.7} & +10.7 \\
9.65 & 45.9 & 45.3 & 45.4 & 43.2 & \best{58.3} & 44.3 & 49.8 & +12.4 \\
10.65 & 45.3 & 44.6 & 45.1 & 42.9 & 44.3 & 44.2 & \best{49.8} & +4.5 \\
\bottomrule
\end{tabular}
}
\end{table*}

\begin{table*}[t]
\centering
\small
\caption{\textbf{Extreme compression below 10M parameters} on BERT\textsubscript{base} (\(\sim\)11.5\(\times\); word-embedding factorized ALBERT-style at rank 64, remaining budget at \(c \approx 13.3\)). Four-task average and two representative tasks. CoLA collapses to 0 for every method at this rate. \textit{Takeaway:} at the extreme, aggressive block allocation wins decisively (+10.8 average, driven by STS-B).}
\label{tab:bert-10m}
\fitbox
{
\begin{tabular}{lrrr}
\toprule
\textbf{Allocation} & \textbf{4-task Avg} & \textbf{STS-B (PCC)} & \textbf{SST-2 (Acc)} \\
\midrule
uniform & 44.1 & \(\sim\)13 & \ph \\
DTFMW-LP (block) & \best{54.9} & \best{54.9} & \best{83.6} \\
\bottomrule
\end{tabular}
}
\end{table*}

\begin{table*}[!t]
\centering
\small
\caption{\textbf{Wanda sparsity on BERT\textsubscript{base}} (primary metric per task: CoLA MCC, MRPC F1, SST-2 accuracy, STS-B Pearson). DTFMW uses the per-weight saliency curve with a density floor. Dense fine-tuned baselines: CoLA 0.55, MRPC 0.87, SST-2 0.93, STS-B 0.88.}
\label{tab:wanda-bert}
\fitbox
{
\begin{tabular}{llrrrrr@{\hspace{18pt}}rrrrr}
\toprule
& & \multicolumn{5}{c}{\textbf{One-shot (no fine-tuning)}} & \multicolumn{5}{c}{\textbf{After fine-tuning (mask preserved)}} \\
\cmidrule(lr){3-7} \cmidrule(lr){8-12}
\textbf{Task} & \textbf{kept} & uniform & DIFW\textsubscript{F} (L) & DIFW\textsubscript{F} (B) & DTFMW (L) & DTFMW (B) & uniform & DIFW\textsubscript{F} (L) & DIFW\textsubscript{F} (B) & DTFMW (L) & DTFMW (B) \\
\midrule
CoLA & 0.5 & 0.533 & 0.535 & 0.539 & 0.253 & \best{0.546} & \best{0.552} & 0.545 & 0.537 & 0.516 & 0.535 \\
CoLA & 0.3 & 0.170 & 0.151 & 0.188 & 0.024 & \best{0.207} & 0.472 & 0.458 & 0.467 & 0.135 & \best{0.478} \\
CoLA & 0.1 & 0.000 & 0.000 & 0.000 & 0.000 & 0.000 & 0.000 & 0.000 & 0.000 & 0.000 & 0.000 \\
\midrule
MRPC & 0.5 & 0.875 & \best{0.880} & 0.874 & 0.584 & 0.828 & 0.873 & 0.878 & 0.878 & \best{0.888} & 0.878 \\
MRPC & 0.3 & 0.007 & 0.021 & 0.035 & \best{0.808} & 0.000 & 0.869 & 0.865 & \best{0.870} & 0.827 & 0.859 \\
MRPC & 0.1 & 0.812 & 0.812 & 0.812 & 0.812 & 0.812 & 0.812 & 0.812 & 0.812 & 0.812 & 0.812 \\
\midrule
SST-2 & 0.5 & 0.917 & 0.919 & 0.920 & 0.811 & \best{0.922} & \best{0.927} & 0.922 & \best{0.927} & 0.919 & 0.924 \\
SST-2 & 0.3 & 0.740 & \best{0.761} & 0.758 & 0.743 & 0.701 & \best{0.914} & 0.909 & 0.912 & 0.891 & 0.912 \\
SST-2 & 0.1 & 0.509 & 0.509 & 0.509 & 0.509 & 0.509 & 0.815 & 0.818 & \best{0.825} & 0.821 & 0.818 \\
\midrule
STS-B & 0.5 & 0.869 & 0.868 & \best{0.870} & 0.843 & 0.864 & \best{0.881} & \best{0.881} & 0.880 & 0.879 & \best{0.881} \\
STS-B & 0.3 & 0.642 & 0.633 & 0.667 & 0.170 & \best{0.697} & 0.866 & 0.866 & 0.869 & 0.858 & \best{0.873} \\
STS-B & 0.1 & \(-\)0.053 & \(-\)0.057 & \(-\)0.060 & \best{0.060} & \(-\)0.056 & 0.106 & 0.108 & 0.106 & \best{0.222} & 0.141 \\
\midrule
\textbf{Avg} & 0.5 & 0.799 & 0.800 & \best{0.800} & 0.623 & 0.790 & \best{0.808} & 0.806 & 0.805 & 0.800 & 0.805 \\
\textbf{Avg} & 0.3 & 0.390 & 0.392 & 0.412 & \best{0.436} & 0.401 & \best{0.780} & 0.775 & 0.779 & 0.678 & \best{0.780} \\
\textbf{Avg} & 0.1 & 0.317 & 0.316 & 0.315 & \best{0.345} & 0.316 & 0.433 & 0.434 & 0.436 & \best{0.464} & 0.443 \\
\bottomrule
\end{tabular}
}
\end{table*}

Table~\ref{tab:rate-sweep} sweeps the global compression rate \( c \) from 1.63 to 10.65 on BERT\textsubscript{base} with the FWSVD base (average primary metric over MRPC, STS-B, CoLA, and SST-2, after fine-tuning).
Three regimes emerge, and the value of allocation grows with compression aggressiveness:
\begin{itemize}[nosep]
    \item \( c < 2 \): uniform allocation is already near-lossless; there is nothing to reallocate.
    \item \( c \approx 2\text{--}3.65 \): DIFW\textsubscript{Fair} and DTFMW (block) lead by +1.5 to +2.5 points.
    \item \( c \approx 4.65\text{--}9.65 \): block-granularity knapsack variants (DTFMW and DTFMW-LP) dominate by +9.0 to +13.7 points (peak at \( c = 7.65 \), reached by DTFMW-LP); at \( c = 10.65 \) a saturation floor (\(\sim\)45) is reached, which DTFMW (block) still breaks by +4.5.
\end{itemize}
Layer-granularity DTFMW variants are unstable at every rate, confirming that the extremal allocation needs block structure.
At an extreme \(\sim\)11.5\(\times\) budget (below 10M parameters, with the word-embedding matrix factorized ALBERT-style at rank 64), DTFMW-LP (block) reaches a four-task average of 54.9 vs.\ 44.1 for uniform (+10.8), driven by STS-B (54.9 Pearson correlation vs.\ \(\sim\)13 for all alternatives); CoLA collapses to 0 for every method at this rate (Table~\ref{tab:bert-10m}).

\FloatBarrier
\section{Wanda Sparsity on BERT}
\label{sec:appendix-wanda-bert}
Table~\ref{tab:wanda-bert} applies the allocations to Wanda sparsity on BERT\textsubscript{base} across four GLUE tasks and three kept fractions, one-shot and after mask-preserved fine-tuning.
One-shot, block-granularity allocation beats uniform on most tasks at kept 0.5/0.3 (e.g., STS-B at kept 0.3: DTFMW (block) 0.697 vs.\ uniform 0.642), while DTFMW (layer) collapses --- the same block-over-layer theme as in the low-rank experiments.
Fine-tuning washes the gaps out on this small model: at kept 0.5/0.3 all allocations land within noise, consistent with allocation being a one-shot phenomenon at mild rates.
At kept 0.1 (90\% sparsity) all methods collapse.

\FloatBarrier
\section{Additional Experimental Results}
\label{sec:appendix-2}

\subsection{BERT Experimental Setup}
\label{sec:appendix-bert-setup}

\begin{figure}[!htbp]
\centering
\includegraphics[width=0.95\linewidth]{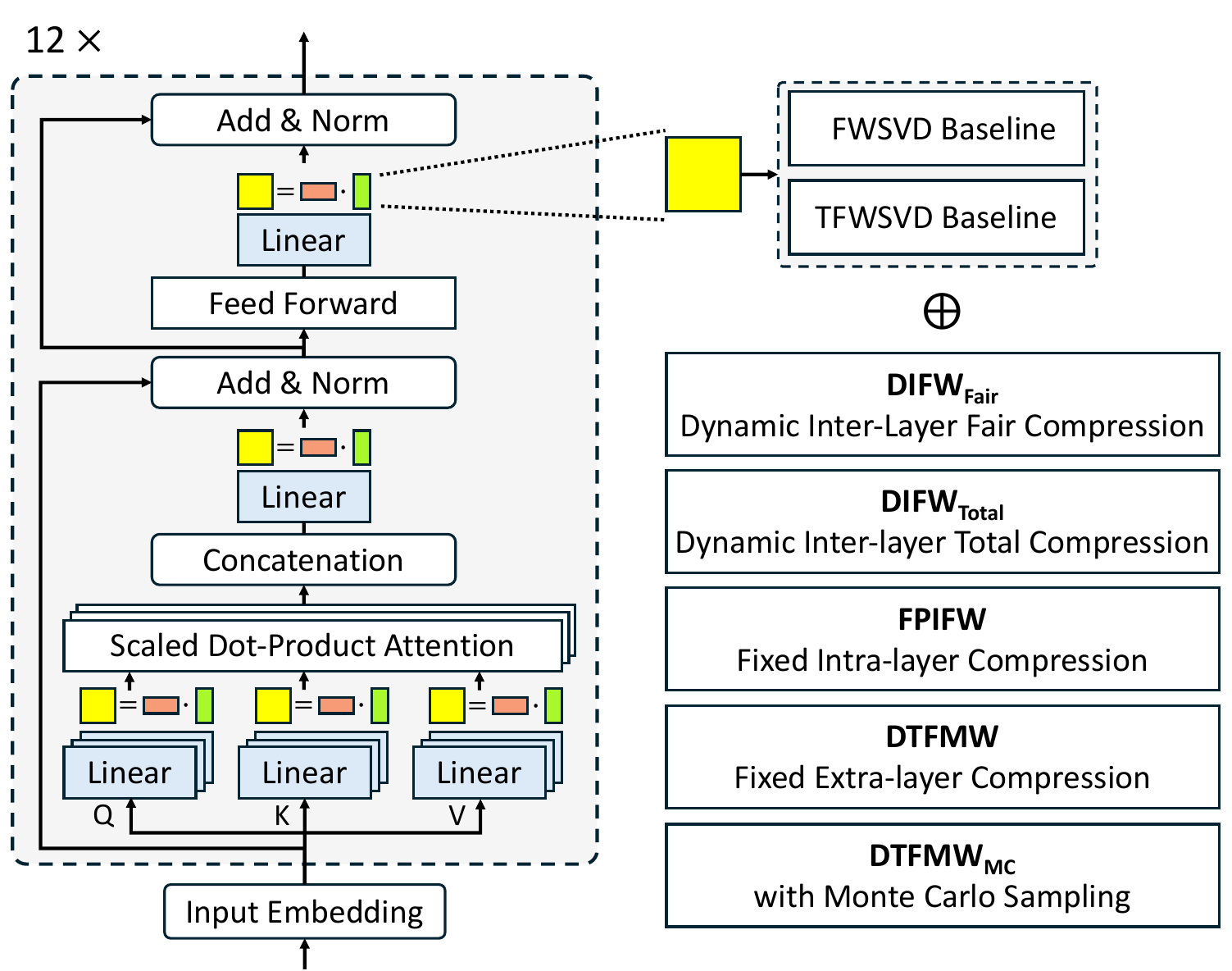}
    \caption{\textbf{Overview of the BERT experimental setup.}
    Linear layers within the 12-layer Transformer encoder are subject to compression.}
    \label{fig:bertcompresslayers}
\end{figure}

\noindent \textbf{Language tasks and datasets.}
We evaluate the compression methods on the GLUE benchmark~\cite{wang2018glue}, a widely used suite for assessing pretrained language models under compression~\cite{khetan2020schubert}.
Our evaluation covers seven tasks across three categories: classification, similarity, and inference.
The single-sentence classification tasks include CoLA~\cite{warstadt2019neural} (evaluated using Matthews correlation~\cite{matthews1975comparison}) and SST-2~\cite{socher2013recursive} (accuracy).
The sentence similarity and paraphrase tasks comprise MRPC~\cite{dolan2005automatically} and QQP~\cite{quoraQQP} (both using F1 score), and STS-B~\cite{cer2017semeval} (evaluated using Pearson correlation).
For natural language inference, we use MNLI~\cite{williams2017broad} (accuracy, averaged over matched and mismatched sets) and QNLI~\cite{rajpurkar2016squad} (accuracy).

\vspace{5pt}
\noindent \textbf{Base model and layers to compress.}
We use the standard 12-layer BERT~\cite{devlin2019bert} as the base model.
BERT is a widely adopted benchmark for evaluating model compression due to its strong performance across diverse natural language tasks~\cite{devlin2019bert, choi2021evaluation} and its prevalence in prior compression studies~\cite{khetan2020schubert, wang2022deep, sanh2020movement, cao2020deformer}.
All linear layers are subjected to low-rank compression using the allocation methods of Section~\ref{sec:Method} or the variants of Appendix~\ref{sec:appendix-variants}.
Following \citet{hsulanguage}, the embedding layer is left uncompressed for consistency with related work.
As shown in Figure~\ref{fig:bertcompresslayers}, compression is applied to the query, key, value, and output projection layers in the attention mechanism, as well as the intermediate and output linear layers within each of the 12 Transformer encoder blocks; task-specific heads are not compressed.

\vspace{5pt}
\noindent \textbf{Training and compression setup.}
To establish an uncompressed performance baseline, we fine-tune the standard BERT\textsubscript{base}~\cite{devlin2019bert} without any compression.
For each task, the model is fine-tuned for 3 epochs using a batch size of 32, a learning rate of 2e-5, and a maximum sequence length of 512 tokens.
All GLUE tasks~\cite{wang2018glue} are used to evaluate the compression strategies under the FWSVD~\cite{hsulanguage} and TFWSVD~\cite{hua2022numerical} frameworks, based on our implementation.
To enable a fair comparison, we adopt the same target total compression rate as in the original FWSVD configuration~\cite{hsulanguage}.
For the DTFMW-MC variant, the compression constraint is enforced via a Lagrangian penalty term; since this yields a soft constraint, the final compression ratio is allowed a tolerance of up to \( c \pm 0.3\% \), using the Adam optimizer~\cite{kingma2014adam} (learning rate 0.01, 1000 steps) and 100{,}000 Monte Carlo samples (Appendix~\ref{sec:appendix-1}).
\subsection{GLUE Results at Compression Rate 1.63}
\label{sec:appendix-glue-results}

\begin{table*}[!t]
\centering
\small
\caption{\textbf{GLUE benchmark results using BERT\textsubscript{base} model with compression rate of 1.63.} G-Avg denotes the average performance across all tasks. ``ft.'' denotes the application of post-compression fine-tuning. A score of 0 indicates complete task failure.
The values in parentheses indicate the performance change compared to the base method.
The best result is emphasized using \best{bold}. The results on GLUE benchmark of other compression method with same number of parameters including distilBERT \cite{sanh2019distilbert}, miniLMv2 \cite{wang2021minilmv2} and TinyBERT6 \cite{jiao2020tinybert} are also presented.
}
\label{Tab:GLUE-BERT-66M}
\fitbox
{
\begin{tabular}{lrrrrrrrrl}
\toprule
\textbf{Approach} & \textbf{\#Param} & \textbf{CoLA} & \textbf{MNLI} & \textbf{MRPC} & \textbf{QNLI} & \textbf{QQP} & \textbf{SST-2} & \textbf{STS-B} & \textbf{G-Avg} \\
 & & (MCC) & (Acc) & (F1) & (Acc) & (F1) & (Acc) & (PCC) & \\
\midrule
BERT\(_{\text{base}}\) \cite{devlin2019bert}  & 108.3M & 56.00 & 84.67 & 89.04 & 91.32 & 87.36 & 92.55 & 88.31 & 84.25 \\
\midrule

distilBERT \cite{sanh2019distilbert} & 67.00M & 49.80 & 82.20 & 88.70 & 89.30 & 86.70 & 90.40 & 86.10 & 81.90 \\

miniLMv2 \cite{wang2021minilmv2} & 67.00M & 43.30 & 84.00 & 89.10 & 90.60 & 86.70 & 91.40 &  88.10 & 81.90 \\

TinyBERT6 \cite{jiao2020tinybert} & 67.00M & 41.20 & 83.90 & 90.60 & 90.60 & 87.00 & 92.10 & 89.40 & 82.10 \\

\midrule

\multicolumn{10}{c}{\textit{FWSVD-based methods}} \\
\midrule
FWSVD  (Base) \cite{hsulanguage} & 66.55M & 2.07 & 49.12 & 0.00 & 49.46 & 63.05 & \best{81.07} & 43.40 & 41.17 \\
\hspace{5pt}  w DIFW\textsubscript{Fair} & 66.55M & 6.15 & \best{53.56} & 0.00 & 49.46 & \best{65.07} & 78.90 & \best{47.59} & \best{42.96} \scriptsize $(+1.79)$ \\
\hspace{5pt} w DIFW\textsubscript{Total} & 66.55M & 6.92 & 53.54 & 0.00 & 49.50 & 64.66 & 80.85 & 44.18 & 42.81 \scriptsize $(+1.64)$ \\

\hspace{5pt} w FPIFW & 66.55M & 0.00 & 48.55 & 0.00 & 49.50 & 55.97 & 80.50 & 0.00 & 33.50 \scriptsize $(-7.67)$ \\
\hspace{5pt} w DTFMW-LP & 66.55M & \best{12.90} & 47.94 & \best{0.71} & \best{49.64} & 62.97 & 78.44 & 24.60 & 39.60 \scriptsize $(-1.57)$ \\
\hspace{5pt} w DTFMW-MC & 66.55M & 0.00 & 49.46 & 0.00 & 49.48 & 50.48 & 79.24 & 13.08 & 34.53 \scriptsize $(-6.64)$ \\

\cmidrule(lr){1-10}
FWSVD + ft. (Base) \cite{hsulanguage}  & 66.55M & 48.67 & 83.27 & 89.19 & 89.80 & 87.56 & 90.71 & 87.21 & 82.34 \\
\hspace{5pt}   w DIFW\textsubscript{Fair} + ft. & 66.55M & \best{49.75} & \best{83.45} & 88.47 & 90.13 & 87.48 & \best{91.97} & 87.25 & \best{82.64} \scriptsize $(+0.30)$ \\
\hspace{5pt} w DIFW\textsubscript{Total} + ft. & 66.55M & 48.34 & 83.34 & 88.19 & \best{90.24}  & 87.46 & 91.63 & 87.23 & 82.35 \scriptsize $(+0.01)$ \\
 \hspace{5pt} w FPIFW + ft. & 66.55M & 45.13 & 83.14 & 86.43 & 90.13 & 87.45 & 91.86 & 85.32 & 81.35 \scriptsize $(-0.99)$ \\
\hspace{5pt} w DTFMW-LP + ft. & 66.55M & 46.94 & 83.12 & 88.16 & 90.15 & \best{87.78} & 90.94 & \best{87.54} & 82.09 \scriptsize $(-0.25)$ \\
\hspace{5pt} w DTFMW-MC + ft. & 66.55M & 39.82 & 82.66 & \best{89.24} & 89.05 & 87.75 & 90.37 & 87.44 & 80.90 \scriptsize $(-1.44)$ \\

\midrule
\multicolumn{10}{c}{\textit{TFWSVD-based methods}
} \\
\midrule
 TFWSVD (Base) \cite{hua2022numerical}  & 66.55M & 13.76 & 45.16 & 40.77 & 49.48 & 64.02 & 82.00 & 46.86 & 48.86 \\
  \hspace{5pt} w DIFW\textsubscript{Fair} & 66.55M & 20.55 & 54.40 & 55.00 & 49.48 & 62.96 & 82.34 & \best{48.82} & \best{53.36} \scriptsize $(+4.50)$ \\
\hspace{5pt} w DIFW\textsubscript{Total} & 66.55M & \best{21.02} & \best{54.63} & \best{64.49} & \best{49.83} & 63.20 & \best{83.49} & 34.84 & 53.07 \scriptsize $(+4.21)$ \\
\hspace{5pt} w FPIFW & 66.55M & 0.00 & 48.56 &  0.00 & 49.50 & 55.97 & 80.50 & 0.00 & 33.50 \scriptsize $(-15.36)$ \\
\hspace{5pt} w DTFMW-LP & 66.55M & 20.79 & 53.45 & 20.89 & 49.99 & \best{65.97} & 80.96 & 36.60 & 46.95 \scriptsize $(-1.91)$ \\
\hspace{5pt} w DTFMW-MC & 66.55M & 3.99 & 47.70 & 1.42 & 49.42 & 62.19 & 80.39 & 9.62 & 36.39 \scriptsize $(-12.47)$ \\

\cmidrule(lr){1-10}
 TFWSVD + ft. (Base) \cite{hua2022numerical}  & 66.55M & 46.80 & \best{83.64} & 89.27 & 89.90 & 87.50 & \best{91.86} & 86.98 & 82.28 \\
  \hspace{5pt} w DIFW\textsubscript{Fair} + ft. & 66.55M & 50.04 & 82.95 & 88.66 & \best{90.46} & \best{87.74} & 90.60 & 87.26 & \best{82.53} \scriptsize $(+0.25)$ \\
\hspace{5pt} w DIFW\textsubscript{Total} + ft. & 66.55M & \best{50.06} & 83.09 & 87.30 & 90.28 & 87.57 & 90.60 & 87.37 & 82.32 \scriptsize $(+0.04)$ \\
\hspace{5pt} w FPIFW + ft. & 66.55M & 45.13 & 83.14 & 86.43 & 90.13 & 87.45 & \best{91.86} & 85.32 & 81.35 \scriptsize $(-0.93)$ \\
\hspace{5pt} w DTFMW-LP + ft. & 66.55M & 45.96 & 83.09 & \best{90.43} & 90.06 & 87.57 & 90.02 & \best{87.65} & 82.11 \scriptsize $(-0.17)$ \\
\hspace{5pt} w DTFMW-MC + ft. & 66.55M & 41.43 & 83.01 & 87.54 & 89.20  & 87.14 & 90.37 & 87.13 & 80.83 \scriptsize $(-1.45)$ \\
\bottomrule
\end{tabular}
}
\end{table*}

Table~\ref{Tab:GLUE-BERT-66M} reports the performance of the FI-guided allocation variants on the GLUE benchmark using BERT\textsubscript{base} at a compression rate of 1.63.
The uncompressed BERT\textsubscript{base} model has 108.3M parameters and achieves a G-Avg of 84.25; the compressed models reduce the parameter count to 66.55M while maintaining competitive performance.
Compared to the uniform-allocation FWSVD and TFWSVD baselines, DIFW yields superior performance: without fine-tuning, DIFW\textsubscript{Fair} improves FWSVD from 41.17 to 42.96 and TFWSVD by 4.50 points; with fine-tuning, DIFW\textsubscript{Fair} reaches 82.64, outperforming FWSVD + ft.\ at 82.34.
Under the same compression rate, DIFW\textsubscript{Fair} + ft.\ also outperforms knowledge-distillation-based compressed models including DistilBERT, MiniLMv2, and TinyBERT6.
Tasks such as CoLA, MRPC, and STS-B exhibit high sensitivity to compression when fine-tuning is not applied; fine-tuning enables the compressed models to recover most of the lost performance.


In addition to the pruning method utilizing SVD decomposition, there are several other language model compression techniques, such as knowledge distillation (e.g., MiniLMv2 \cite{wang2021minilmv2}, DistilBERT \cite{sanh2019distilbert}). Experiments have been conducted to evaluate the applicability of the proposed SVD--based methods on these compressed models. The experimental setup follows the same configuration as that used for the BERT model experiments.


Tables \ref{Tab:GLUE-DISTILBERT-43.30M} and \ref{Tab:GLUE-MiniLMv2} present the results of compressing DistilBERT and MiniLMv2, respectively.
It becomes evident that the proposed SVD-based methods can be effectively applied to families of models compressed using knowledge distillation techniques.
Notably, DIFW\textsubscript{Fair} and DIFW\textsubscript{Total}, when applied to FWSVD and TFWSVD, achieved higher G-Avg scores than the baselines in 15 of 16 settings (the exception is FWSVD w DIFW\textsubscript{Fair} on DistilBERT without fine-tuning, \(-0.84\)). Furthermore, the successful application of SVD to already compressed models opens a new avenue for combining compression techniques to achieve the highest compression ratios while still retaining a reasonable level of accuracy.

\begin{table*}[!t]
\centering
\small
\caption{\textbf{GLUE benchmark results using DistilBERT\textsubscript{base} model with compression rate {1.55}.} 
G-Avg denotes the average performance across all tasks. ``ft.'' denotes the application of post-compression fine-tuning. A score of 0 indicates complete task failure. The best result is emphasized using \best{bold}.
}
\label{Tab:GLUE-DISTILBERT-43.30M}
\fitbox
{
\begin{tabular}{lrrrrrrrrl}
\toprule
\textbf{Approach} & \textbf{\#Param} & \textbf{CoLA} & \textbf{MNLI} & \textbf{MRPC} & \textbf{QNLI} & \textbf{QQP} & \textbf{SST-2} & \textbf{STS-B} & \textbf{G-Avg} \\
 & & (MCC) & (Acc) & (F1) & (Acc) & (F1) & (Acc) & (PCC) & \\
\midrule
DistilBERT\(_{\text{base}}\) \cite{sanh2019distilbert}  & 67.00M & 56.00 & 84.67 & 89.04 & 91.32 & 87.36 & 92.55 & 88.31 & 84.25 \\
\midrule
\multicolumn{10}{c}{\textit{FWSVD-based methods}} \\
\midrule
 FWSVD  (Base) \cite{hsulanguage} & 43.30M & 10.24 &	44.22&	75.23&	59.77&	68.55&	79.59&	\best{64.82} &	57.49 \\
\hspace{5pt} w DIFW\textsubscript{Fair} & 43.30M & 13.59 &	41.82 &	82.19 &	62.99 &	68.20 &	79.01 &	48.75 &	56.65 \scriptsize $(-0.84)$ \\
 \hspace{5pt} w DIFW\textsubscript{Total} & 43.30M & \best{14.98} &	43.76 &	\best{82.26} &	64.27 &	\best{70.97} &	79.47 &	56.84 &	\best{58.94} \scriptsize $(+1.45)$   \\
\hspace{5pt} w FPIFW & 43.30M & 1.80 &	42.53 &	5.54 &	50.58 &	67.14 &	79.36 &	38.40 &	40.76 \scriptsize $(-16.73)$ \\
\hspace{5pt} w DTFMW-LP & 43.30M & 3.52 &	\best{44.18} &	81.13 &	55.98 &	68.21 &	79.70 &	27.70 &	51.49 \scriptsize $(-6.00)$ \\
\hspace{5pt} w DTFMW-MC & 43.30M & 9.06 &	38.83 &	80.06 &	\best{66.32} &	69.30 &	\best{80.96} & 38.43 &	50.33 \scriptsize $(-7.16)$ \\
\cmidrule(lr){1-10}
FWSVD + ft. (Base) \cite{hsulanguage}  & 43.30M & 37.92 & 	81.18& 	\best{88.73} &	\best{87.83}  &	86.55 &	89.33 &	85.73 &	79.61 \\
\hspace{5pt} w DIFW\textsubscript{Fair} + ft. & 43.30M & \best{40.82} &	81.31 &	87.52 &	87.70 &	86.49 &	89.45 &	86.21 &	79.93 \scriptsize $(+0.32)$ \\
\hspace{5pt} w DIFW\textsubscript{Total} + ft. & 43.30M & 39.33 &	81.28 &	88.35 &	87.68 &	86.51 &	\best{90.25} &	\best{86.24} &	\best{79.95} \scriptsize $(+0.34)$ \\
 \hspace{5pt} w FPIFW + ft. & 43.30M & 36.89 &	\best{81.36} &	86.51 &	86.64 &	\best{86.63} &	89.11 &	86.58 &	79.10 \scriptsize $(-0.51)$ \\
\hspace{5pt} w DTFMW-LP + ft. & 43.30M & 33.92 &	81.27 &	84.91 &	87.13 &	86.34 &	89.56 &	84.30 &	78.20 \scriptsize $(-1.41)$ \\
\hspace{5pt} w DTFMW-MC + ft. & 43.30M & 34.62 &	80.97 &	86.21 &	87.11 &	86.56 &	90.14 &	85.52 &	78.73 \scriptsize $(-0.88)$ \\

\midrule
\multicolumn{10}{c}{\textit{TFWSVD-based methods}} \\
\midrule
TFWSVD (Base) \cite{hua2022numerical}  & 43.30M & 12.24 & 	42.30 & 80.67 & 57.37 & 63.67 &	\best{83.26} & 	14.13 & 50.52 \\
\hspace{5pt} w DIFW\textsubscript{Fair} & 43.30M & 15.32 &	39.79 &	\best{82.86} &	\best{68.41} &	63.22 &	81.54 &	29.52 &	54.38 \scriptsize $(+3.86)$ \\
   \hspace{5pt} w DIFW\textsubscript{Total} & 43.30M & 16.90 &	40.95 &	82.33 &	66.23 &	66.61 &	81.88 &	\best{44.38} &	\best{57.04} \scriptsize $(+6.52)$ \\
\hspace{5pt} w FPIFW & 43.30M & 5.29 &	43.38 &	30.86 &	61.54 &	62.26 &	81.77 &	0.00 &	40.73 \scriptsize $(-9.79)$ \\
\hspace{5pt} w DTFMW-LP & 43.30M & 11.63 &	42.84 &	80.13 &	66.30 &	67.33 &	79.82 &	0.00 &	49.72 \scriptsize $(-0.80)$ \\
\hspace{5pt} w DTFMW-MC & 43.30M & \best{18.67} &	\best{43.85} &	77.76 &	60.26 &	\best{67.44} &	81.54 &	18.00 &	52.50 \scriptsize $(+1.98)$ \\

\cmidrule(lr){1-10}
TFWSVD + ft. (Base)  \cite{hua2022numerical}  & 43.30M & 36.34 &	80.14 &	87.59 &	86.66 &	85.72 &	89.56 &	85.11 &	78.73 \\
 \hspace{5pt} w DIFW\textsubscript{Fair} + ft. & 43.30M & 37.97 &	\best{81.01} &	\best{87.65} &	\best{87.24} &	\best{86.71} &	\best{90.14} &	\best{85.70} &	\best{79.49} \scriptsize $(+0.76)$ \\
\hspace{5pt} w DIFW\textsubscript{Total} + ft. & 43.30M & \best{39.23} &	\best{81.01} &	87.26 &	86.89 &	86.64 &	89.56 &	85.61 &	79.46 \scriptsize $(+0.73)$ \\
\hspace{5pt} w FPIFW + ft. & 43.30M & 34.84 &	80.79 &	86.47 &	86.69 &	86.60 &	88.53 &	85.49 &	78.49 \scriptsize $(-0.24)$ \\
\hspace{5pt} w DTFMW-LP + ft. & 43.30M & 34.71 &	80.30 &	84.43 &	86.66 &	86.24 &	89.33 &	83.70 &	77.91 \scriptsize $(-0.82)$ \\
\hspace{5pt} w DTFMW-MC + ft. & 43.30M  & 36.53 &	80.98 &	87.06 &	87.02 &	86.46 &	90.25 &	84.96 &	79.04 \scriptsize $(+0.31)$  \\

\bottomrule
\end{tabular}
}
\end{table*}

\begin{table*}[!t]
\centering
\small
\caption{\textbf{GLUE benchmark results using MiniLM-v2 model with compression rate {1.35}.} 
G-Avg denotes the average performance across all tasks. ``ft.'' denotes the application of post-compression fine-tuning. A score of 0 indicates complete task failure. The best result is emphasized using \best{bold}.
}
\label{Tab:GLUE-MiniLMv2}
\fitbox
{
\begin{tabular}{lrrrrrrrrl}
\toprule
\textbf{Approach} & \textbf{\#Param} & \textbf{CoLA} & \textbf{MNLI} & \textbf{MRPC} & \textbf{QNLI} & \textbf{QQP} & \textbf{SST-2} & \textbf{STS-B} & \textbf{G-Avg} \\
 & & (MCC) & (Acc) & (F1) & (Acc) & (F1) & (Acc) & (PCC) & \\
\midrule
MiniLM-v2 \cite{wang2021minilmv2} & 22.71M & 14.58 &	79.25 &	82.65 &	87.42 &	86.14 &	89.11 &	80.83 &	74.28 \\
\midrule
\multicolumn{10}{c}{\textit{FWSVD-based methods}} \\
\midrule
 FWSVD  (Base) \cite{hsulanguage} & 16.80M  & 1.05 &	42.01 &	74.93 &	49.61 &	46.29 &	72.29 &	9.20 &	42.20  \\
 \hspace{5pt} w DIFW\textsubscript{Fair} & 16.80M & 0.00 &	43.28 &	74.60 &	52.50 &	51.76 &	74.43 &	12.30 &	\best{44.12} \scriptsize $(+1.92)$  \\
\hspace{5pt} w DIFW\textsubscript{Total} & 16.80M & 0.00 &	39.31 &	74.64 &	54.88 &	50.14 &	71.22 &	9.03 &	42.75 \scriptsize $(+0.75)$  \\
\hspace{5pt} w FPIFW & 16.80M & 2.00 &	43.45 &	2.13 &	49.73 &	42.27 &	\best{77.52} &	7.16 &	32.04 \scriptsize $(-10.16)$  \\
\hspace{5pt} w DTFMW-LP & 16.80M& 0.00 &	34.17 &	\best{81.46} &	\best{57.84} &	3.85 &	76.03 &	0.00 &	36.19 \scriptsize $(-6.01)$   \\
\hspace{5pt} w DTFMW-MC & 16.80M & \best{5.59} &	\best{43.65} &	42.93 &	56.58 &	\best{54.96} &	72.94 &	\best{21.73} &	42.63 \scriptsize $(+0.43)$  \\
\cmidrule(lr){1-10}
FWSVD + ft. (Base) \cite{hsulanguage}   & 16.80M & 11.25 &	78.06 &	81.32 &	85.45 &	84.30 &	88.42 &	76.92 &	72.24 \\
\hspace{5pt} w DIFW\textsubscript{Fair} + ft. & 16.80M  & 11.13 &	\best{79.15}  &	82.13 &	\best{86.40} &	85.06 &	\best{88.65} &	76.99 &	72.79 \scriptsize $(+0.55)$   \\
\hspace{5pt} w DIFW\textsubscript{Total} + ft. & 16.80M & \best{12.82} &	78.96 &	82.20 &	\best{86.40} &	85.16 &	88.42 &	\best{77.58} &	\best{73.08}  \scriptsize $(+0.84)$  \\
 \hspace{5pt} w FPIFW + ft. & 16.80M & 9.88 &	79.00 &	82.22 &	85.70 &	\best{85.17} &	87.96 &	77.37 &	72.47 \scriptsize $(+0.23)$ \\
\hspace{5pt} w DTFMW-LP + ft. & 16.80M & 7.13 &	78.20 &	\best{83.54} &	85.70 &	84.40 &	87.61 &	71.72 &	71.19 \scriptsize $(-1.05)$  \\
\hspace{5pt} w DTFMW-MC + ft. & 16.80M & 6.64 &	78.53 &	82.97 &	84.70 &	84.54 &	87.27 &	75.76 &	71.49 \scriptsize $(-0.75)$  \\

\midrule
\multicolumn{10}{c}{\textit{TFWSVD-based methods}} \\
\midrule
TFWSVD (Base) \cite{hua2022numerical}  & 16.80M & \best{2.00} &	41.45 &	2.13 &	49.73 &	42.27 &	75.52 &	7.16 &	31.47 \\
  \hspace{5pt} w DIFW\textsubscript{Fair} & 16.80M & 0.00 &	\best{46.54} &	80.00 &	50.34 &	39.50 &	77.52 & 16.60 & \best{44.36} \scriptsize $(+12.89)$  \\
\hspace{5pt} w DIFW\textsubscript{Total} & 16.80M & 0.00 &	44.28 &	65.31 &	50.17 &	38.71 &	77.29 &	15.42 &	41.60 \scriptsize $(+10.13)$  \\
\hspace{5pt} w FPIFW & 16.80M & 0.00 &	46.25 &	22.91 &	49.48 &	42.62 &	74.89 &  0.00 & 33.73 \scriptsize $(+2.26)$  \\
\hspace{5pt} w DTFMW-LP & 16.80M & 0.00 &	39.65 &	\best{81.58} &	56.98 &	5.88 &	77.18 &	4.41 &	37.95 \scriptsize $(-6.48)$   \\
\hspace{5pt} w DTFMW-MC & 16.80M & 0.00 &	49.68 &	36.96 &	\best{58.65} &	\best{46.08} &	\best{78.90} &	\best{23.49} &	41.97 \scriptsize $(+10.5)$  \\

\cmidrule(lr){1-10}
TFWSVD + ft. (Base) \cite{hua2022numerical} & 16.80M & 4.88 &	78.00 &	81.22 &	85.70 &	85.17 &	87.96 &	\best{77.37} &	71.47 \\
 \hspace{5pt} w DIFW\textsubscript{Fair} + ft. & 16.80M & 4.43 &	78.75 &	82.24 &	\best{86.09} &	85.17 &	\best{89.56} &	75.38 &	71.66 \scriptsize $(+0.19)$  \\
 \hspace{5pt} w DIFW\textsubscript{Total} + ft. & 16.80M & 5.22 &	78.75 &	\best{83.20} &	85.81 &	85.16 &	88.99 & 75.46 &	\best{71.80} \scriptsize $(+0.33)$ \\
\hspace{5pt} w FPIFW + ft. & 16.80M & 3.84 &	\best{78.77} &	80.72 &	85.48 &	\best{85.20} &	88.99 & 66.62 &	69.95 \scriptsize $(-1.52)$  \\
\hspace{5pt} w DTFMW-LP + ft. & 16.80M & \best{5.71} &	77.71 &	81.54 &	85.19 &	84.45 &	88.88 & 67.22 &	70.10 \scriptsize $(-1.37)$  \\
\hspace{5pt} w DTFMW-MC + ft. & 16.80M & 3.71  &	78.38 &	81.96 &	84.42 &	84.64 &	88.42 &	67.94 & 69.92 \scriptsize $(-1.55)$  \\

\bottomrule
\end{tabular}
}
\end{table*}

\subsection{Rank--Information Ablation on CoLA}
\label{sec:appendix-rank-info}
\begin{figure*}[!t]
    \begin{subfigure}[b]{0.49\textwidth}
        \centering
        \includegraphics[width=\textwidth]{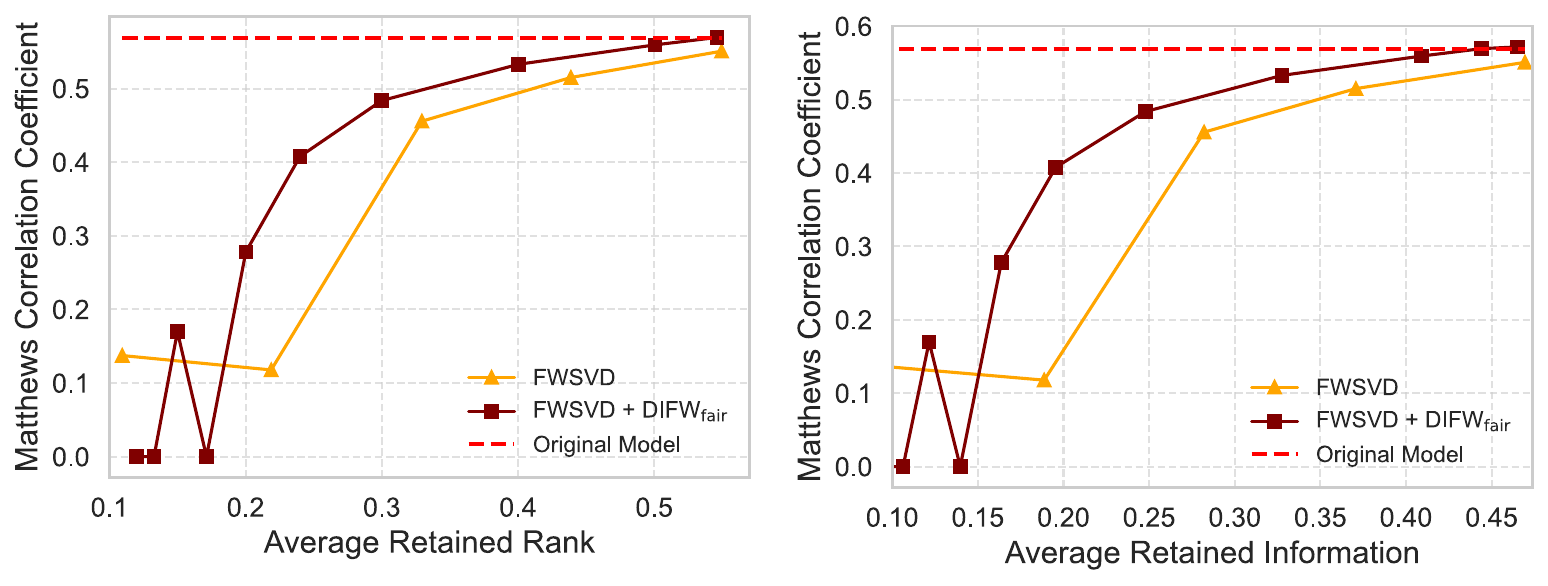}
        \caption{Comparison between FWSVD and FWSVD + DIFW\textsubscript{Fair}.}
        \label{fig:fwsvd_with_difw_fair_a}
    \end{subfigure}
    \hfill
    \begin{subfigure}[b]{0.49\textwidth}
        \centering
        \includegraphics[width=\textwidth]{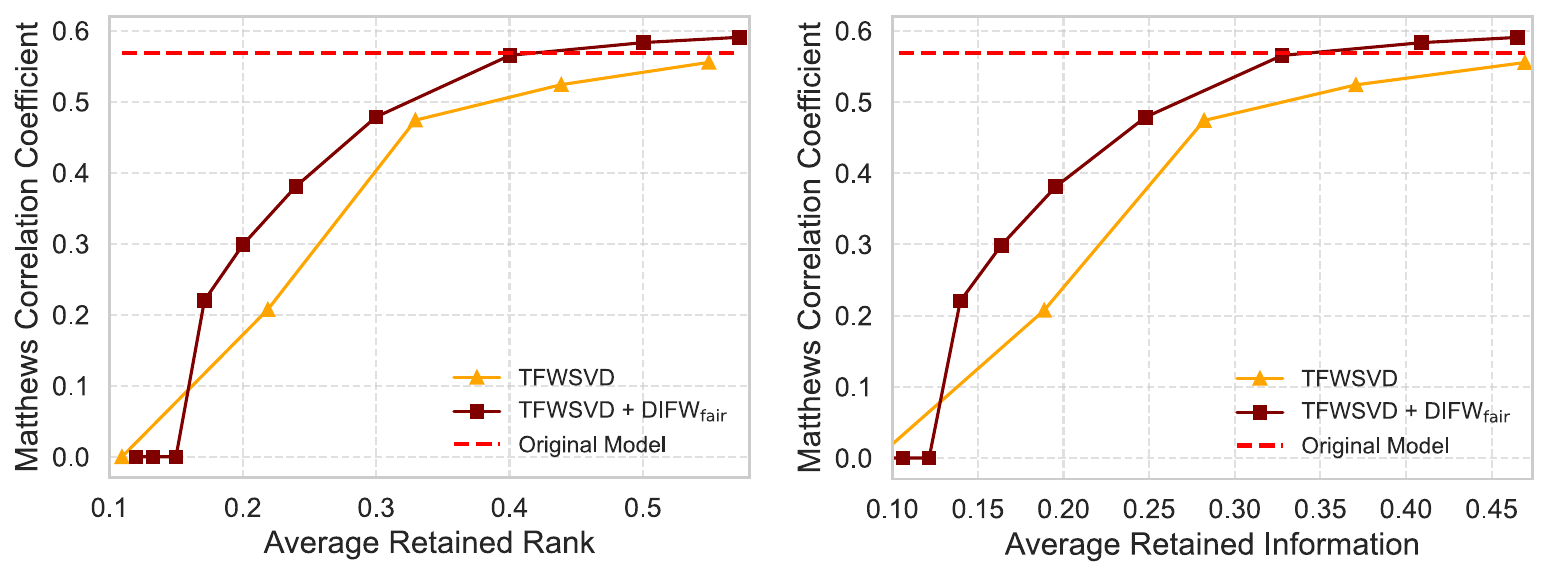}
        \caption{Comparison between TFWSVD and TFWSVD + DIFW\textsubscript{Fair}.}
        \label{fig:fwsvd_with_difw_fair_b}
    \end{subfigure}
    \hfill
    \begin{subfigure}[b]{0.49\textwidth}
        \centering
        \includegraphics[width=\textwidth]{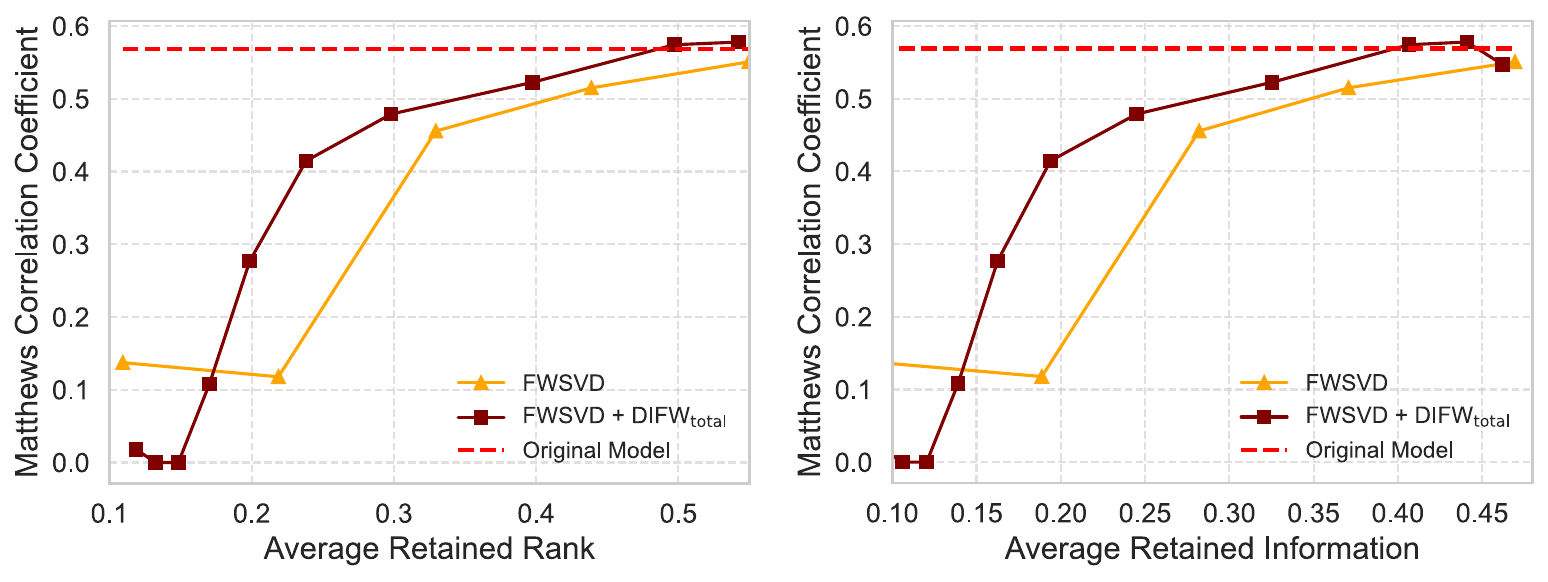}
        \caption{Comparison between FWSVD and FWSVD + DIFW\textsubscript{Total}.}
        \label{fig:fwsvd_with_difw_c}
    \end{subfigure}
    \hfill
    \begin{subfigure}[b]{0.49\textwidth}
        \centering
        \includegraphics[width=\textwidth]{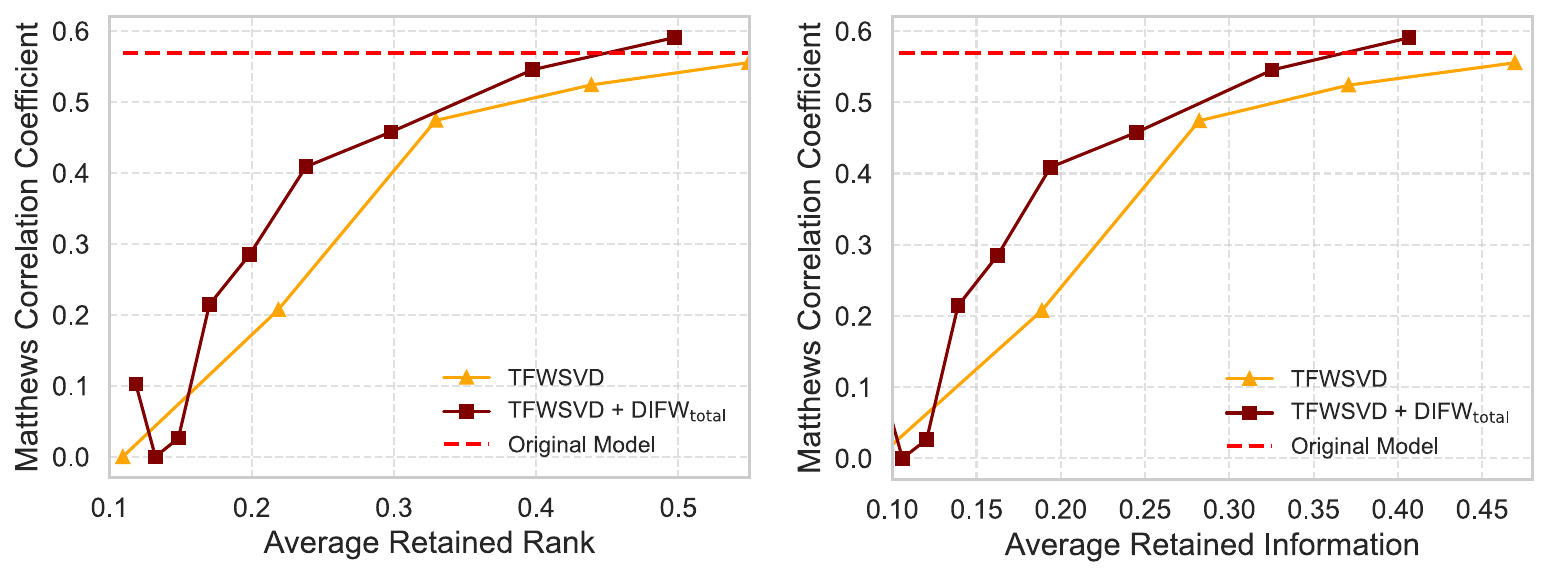}
        \caption{Comparison between TFWSVD and TFWSVD + DIFW\textsubscript{Total}.}
        \label{fig:fwsvd_with_difw_d}
    \end{subfigure}
    \caption{Effect of average retained rank and retained information on Matthews Correlation Coefficient (MCC), evaluated on the CoLA task using BERT\textsubscript{base}.}
    \label{fig:fwsvd_with_difw_fair}
\end{figure*}

We analyze the behavior of the FI-guided allocation under varying levels of retained rank and retained information.
Figure~\ref{fig:fwsvd_with_difw_fair} demonstrates that DIFW\textsubscript{Fair} and DIFW\textsubscript{Total} consistently outperform their respective baselines, FWSVD and TFWSVD, across nearly all compression levels: for the same average retained rank or retained information, the FI-guided variants achieve higher Matthews Correlation Coefficient (MCC) scores on CoLA.
The performance gap is particularly significant at lower ranks, where the baseline methods tend to fail or degrade sharply, while the FI-guided variants maintain reasonable accuracy.

\begin{table*}[t]
\centering
\small
\caption{\textbf{Wanda sparsity on LLaMA-3-8B}, one-shot, per kept fraction. DTFMW floored at \(0.5\cdot\)kept. \textit{Takeaway:} LLaMA-3 tolerates sparsity well (kept 0.6 near-dense); DIFW \(\approx\) uniform at every rate (seed-confirmed tie at 0.4), the floored DTFMW is best at kept 0.8 but degrades with aggressiveness, and the searched/proportional profiles degrade at every rate.}
\label{tab:wanda-llama3}
\fitbox
{
\begin{tabular}{lrrrr}
\toprule
\textbf{Allocation} & \textbf{kept 0.2} & \textbf{kept 0.4} & \textbf{kept 0.6} & \textbf{kept 0.8} \\
\midrule
uniform & 679.9 / 29.7 & 28.55 / 44.6 & \best{9.38} / 57.4 & 8.30 / 60.0 \\
DIFW\textsubscript{Fair} (layer) & 709.4 / 30.0 & 28.90 / 44.6 & \best{9.38} / 57.3 & 8.31 / 59.9 \\
DIFW\textsubscript{Fair} (block) & \best{659.4} / 29.9 & \best{28.54} / 44.6 & 9.42 / 57.5 & 8.33 / 59.8 \\
DTFMW (layer, floored) & 2658 / 30.2 & 178.3 / 31.4 & 15.78 / 50.6 & 9.00 / 58.2 \\
DTFMW (block, floored) & 2498 / 29.7 & 58.03 / 35.4 & 10.85 / 53.9 & \best{8.25} / 59.8 \\
\midrule
Bolaco (transplant) & 3940 / 29.9 & 83.25 / 35.5 & 14.20 / 51.9 & 9.90 / 56.8 \\
FLRC (native) & 28333 / 29.6 & 426.6 / 30.4 & 42.95 / 42.1 & 13.24 / 52.6 \\
\bottomrule
\end{tabular}
}
\end{table*}

\FloatBarrier
\section{LLaMA-3-8B: Full Per-Rate Results}
\label{sec:appendix:llama3}
Tables~\ref{tab:wanda-llama3}, \ref{tab:svdllm-llama3}, and~\ref{tab:gptq-llama3} give the complete per-rate LLaMA-3 results (with zero-shot) behind the main-text battles (Tables~\ref{tab:svd-battle}--\ref{tab:gptq-battle}). Cells: WikiText-2 PPL\(\downarrow\) / 7-task ZS\(\uparrow\), best PPL per column in bold; ``--'' = zero-shot not evaluated; dense PPL \(\approx\) 8.0. Bolaco is an L2-type-ratio transplant; FLRC uses importance recomputed natively from LLaMA-3 gradients. These per-rate tables report single-calibration (seed 0) runs; the seeded means behind the headline cells (Tables~\ref{tab:svd-battle}--\ref{tab:gptq-battle} and Appendix~\ref{sec:appendix:fullgrid}) therefore differ, substantially so for GPTQ, where calibration variance is large.

\begin{table*}[!htbp]
\centering
\small
\caption{\textbf{Whitened low-rank (our SVD-LLM operator, whitening-only) on LLaMA-3-8B.} \textit{Takeaway:} LLaMA-3 is drastically more brittle to low-rank than LLaMA-2 (kept 0.6: 71--74 vs.\ \(\sim\)16 PPL); DIFW\textsubscript{Fair} leads at 0.4 and 0.8, Bolaco's MLP-protective profile wins the 0.6 cell, and ZS confirms all kept \(\leq\) 0.6 models are heavily damaged.}
\label{tab:svdllm-llama3}
\fitbox
{
\begin{tabular}{lrrr}
\toprule
\textbf{Allocation} & \textbf{kept 0.4} & \textbf{kept 0.6} & \textbf{kept 0.8} \\
\midrule
uniform & 540.5 / 29.8 & 74.38 / 31.0 & 15.50 / 42.8 \\
DIFW\textsubscript{Fair} (layer) & 538.4 / 29.7 & 72.74 / 31.2 & \best{15.29} / 43.0 \\
DIFW\textsubscript{Fair} (block) & \best{536.6} / 29.8 & 70.87 / 31.0 & 15.30 / 43.1 \\
DTFMW (block) & 1141 / 29.5 & 379.5 / 29.7 & 30.89 / 35.7 \\
\midrule
Bolaco (transplant) & 636.9 / 30.0 & \best{60.97} / 33.5 & 16.66 / 46.2 \\
FLRC (native) & 2019 / 30.3 & 182.1 / 30.5 & 30.87 / 35.9 \\
\bottomrule
\end{tabular}
}
\end{table*}

\begin{table*}[!htbp]
\centering
\small
\caption{\textbf{GPTQ on LLaMA-3-8B}, per-layer mixed precision. Unlike LLaMA-2, layer granularity is \emph{not} degenerate (GQA skews per-projection Fisher shares enough to flip integer bits). \textit{Takeaway:} uniform 4-bit collapses (seed-robust, Table~\ref{tab:gptq-battle}) while Fisher mixed-precision stays near-dense; at 3-bit everything is damaged and DIFW (layer) degrades least.}
\label{tab:gptq-llama3}
\fitbox
{
\begin{tabular}{lrrr}
\toprule
\textbf{Allocation} & \textbf{2-bit} & \textbf{3-bit} & \textbf{4-bit} \\
\midrule
uniform & \(4.4\times10^{5}\) / 30.1 & 300.8 / 34.4 & 1042.9 / 51.6 \\
DIFW\textsubscript{Fair} (layer) & --- & \best{46.18} / 41.9 & 13.81 / 56.0 \\
DIFW\textsubscript{Fair} (block) & --- & 166.7 / 39.1 & \best{9.30} / 56.4 \\
DTFMW (layer) & --- & 25652 / -- & 1512 / -- \\
DTFMW (block) & --- & 800.6 / -- & 1150 / -- \\
Bolaco (transplant) & --- & 298.6 / 40.6 & 1340.6 / 49.7 \\
FLRC (native) & --- & 772.4 / 30.0 & 15.28 / 50.2 \\
\bottomrule
\end{tabular}
}
\end{table*}

\begin{table*}[!htbp]
\centering
\small
\caption{\textbf{SVD-LLM battle after recovery} (LLaMA-2; PPL\(\downarrow\)). ``+ update'' = SVD-LLM's closed-form weight update (\(n=4\) seeds where marked, \(\dagger\) = single seed). Cells: PPL / ZS as in Table~\ref{tab:svd-battle}; headers give remaining parameters. \textit{Takeaway:} recovery does not wash out the allocation --- DIFW leads before and after the update.}
\label{tab:svd-recovery}
\fitbox
{
\begin{tabular}{lrrr}
\toprule
& \multicolumn{3}{c}{\textbf{+ update, kept}} \\
\cmidrule(lr){2-4}
\textbf{Allocation} & \textbf{0.4} (2.9B) & \textbf{0.6} (4.1B) & \textbf{0.8} (5.4B) \\
\midrule
\multicolumn{4}{l}{\emph{Baselines}} \\
uniform & 53.20 {\tiny$\pm$1.17} / 30.3 & 16.17 {\tiny$\pm$0.15} / 34.8 & 8.79\(^\dagger\) / 42.1 \\
Bolaco~\cite{ji2024bolaco} & 54.72\(^\dagger\) / 30.0 & 18.32\(^\dagger\) / 35.7 & \best{8.72\(^\dagger\) / 45.8} \\
FLRC~\cite{lu2025flrc} & 117.2\(^\dagger\) / 30.2 & 34.45\(^\dagger\) / 32.3 & 15.31\(^\dagger\) / 37.1 \\
\midrule
\multicolumn{4}{l}{\emph{Ours (Fisher-guided)}} \\
DIFW\textsubscript{Fair} (L) & 50.53\(^\dagger\) / 30.2 & 15.74\(^\dagger\) / 34.9 & 8.75\(^\dagger\) / 42.6 \\
DIFW\textsubscript{Fair} (B) & \best{48.82 {\tiny$\pm$0.86} / 30.3} & \best{15.90 {\tiny$\pm$0.21} / 35.2} & 8.86\(^\dagger\) / 42.4 \\
DTFMW (L / B) & \multicolumn{3}{c}{collapse} \\
\bottomrule
\end{tabular}
}
\end{table*}

\FloatBarrier
\section{Paired Seed Statistics and Zero-Shot Robustness}
\label{sec:appendix:fullgrid}
This section collects the recovery battle (Table~\ref{tab:svd-recovery}), the paired per-seed statistics behind every main-text comparison (Table~\ref{tab:llm-seeds}), and the multi-seed zero-shot versions of the headline cells (Table~\ref{tab:zs-seeds}). The seven zero-shot tasks are ARC-Easy, ARC-Challenge, HellaSwag, OpenBookQA, WinoGrande, MathQA, and PIQA.

Paired per-seed differences (with verdicts) and seed-robust zero-shot accuracy for the headline cells of the main-text battles.

\begin{table*}[!t]
\centering
\small
\caption{\textbf{Seed robustness on LLaMA-2-7B and LLaMA-3-8B} (\(n = 5\) seeds varying Fisher + calibration unless marked \(n = 4\); WikiText-2 PPL\(\downarrow\); paired \(\Delta\) = DIFW\textsubscript{Fair} (block) \(-\) uniform per seed). ``all \(k\)'' = DIFW wins on every seed. This table adds the \emph{paired} per-seed statistics behind the main-text battles (Tables~\ref{tab:svd-battle}--\ref{tab:gptq-battle}). Cells without seeds anywhere: Wanda kept 0.2 and GPTQ 2-bit (all-methods collapse). DIFW's rank rounding keeps 1.5--2.5\% more parameters than uniform on the low-rank base. \textit{Takeaway:} DIFW's advantage is seed-robust exactly where compression bites --- the low-rank base on both backbones (with and without the closed-form update), aggressive sparsity on LLaMA-2, and above all LLaMA-3 quantization, where uniform's perplexity collapses on every seed at 4-bit --- and vanishes into noise where the primitive saturates (mild rates) or where quantization noise dominates (GPTQ 3-bit).}
\label{tab:llm-seeds}
\fitbox
{
\begin{tabular}{lrrrl}
\toprule
\textbf{Cell} & \textbf{uniform} & \textbf{DIFW\textsubscript{Fair} (block)} & \textbf{paired \(\Delta\)} & \textbf{verdict} \\
\midrule
\multicolumn{5}{l}{\emph{LLaMA-2-7B}} \\
SVD-LLM, kept 0.4 (whitening) & 81.52 \(\pm\) 2.10 & \best{72.58 \(\pm\) 1.50} & \(-\)8.944 \(\pm\) 0.882 & \textbf{DIFW, all 5} \\
SVD-LLM, kept 0.4 (+ update, n = 4) & 53.20 \(\pm\) 1.17 & \best{48.82 \(\pm\) 0.86} & \(-\)4.379 \(\pm\) 0.429 & \textbf{DIFW, all 4} \\
SVD-LLM, kept 0.6 (whitening) & 16.05 \(\pm\) 0.15 & \best{15.78 \(\pm\) 0.16} & \(-\)0.273 \(\pm\) 0.095 & \textbf{DIFW, all 5} \\
SVD-LLM, kept 0.6 (+ update, n = 4) & 16.17 \(\pm\) 0.15 & \best{15.90 \(\pm\) 0.21} & \(-\)0.272 \(\pm\) 0.081 & \textbf{DIFW, all 4} \\
SVD-LLM, kept 0.8 (whitening) & \best{8.37 \(\pm\) 0.04} & 8.46 \(\pm\) 0.06 & \(+\)0.092 \(\pm\) 0.036 & uniform (marginal) \\
Wanda, kept 0.4 & 10.80 \(\pm\) 0.07 & \best{10.42 \(\pm\) 0.11} & \(-\)0.387 \(\pm\) 0.058 & \textbf{DIFW, all 5} \\
Wanda, kept 0.6 & \best{6.07 \(\pm\) 0.01} & 6.10 \(\pm\) 0.01 & \(+\)0.035 \(\pm\) 0.008 & uniform (marginal) \\
Wanda, kept 0.8 & \best{5.59 \(\pm\) 0.00} & 5.60 \(\pm\) 0.00 & \(+\)0.015 \(\pm\) 0.002 & tie \\
GPTQ, 4-bit & 6.10 \(\pm\) 0.02 & \best{6.08 \(\pm\) 0.03} & \(-\)0.020 \(\pm\) 0.028 & tie \\
GPTQ, 3-bit & \best{10.59 \(\pm\) 0.72} & 10.49 \(\pm\) 1.07 & \(-\)0.100 \(\pm\) 0.702 & tie (calib.\ noise) \\
\midrule
\multicolumn{5}{l}{\emph{LLaMA-3-8B}} \\
GPTQ, 4-bit & 1496.8 \(\pm\) 797.1 (collapse) & \best{13.02 \(\pm\) 8.11} & \(-\)1483.8 \(\pm\) 793.9 & \textbf{DIFW, all 5 (rescue)} \\
GPTQ, 3-bit & \best{401.6 \(\pm\) 317.7} & 1065.4 \(\pm\) 1576.0 & \(+\)663.7 \(\pm\) 1440.7 & tie (both collapse) \\
SVD-LLM, kept 0.4 & 577.8 \(\pm\) 58.1 & \best{553.7 \(\pm\) 48.4} & \(-\)24.03 \(\pm\) 20.36 & \textbf{DIFW, all 5} \\
SVD-LLM, kept 0.6 & 73.02 \(\pm\) 2.20 & \best{71.34 \(\pm\) 2.13} & \(-\)1.678 \(\pm\) 1.119 & \textbf{DIFW, all 5} \\
SVD-LLM, kept 0.8 & 15.49 \(\pm\) 0.08 & \best{15.33 \(\pm\) 0.13} & \(-\)0.154 \(\pm\) 0.060 & \textbf{DIFW, all 5} \\
Wanda, kept 0.4 & \best{28.46 \(\pm\) 0.20} & 28.60 \(\pm\) 0.40 & \(+\)0.140 \(\pm\) 0.232 & tie \\
Wanda, kept 0.6 & \best{9.39 \(\pm\) 0.02} & 9.42 \(\pm\) 0.01 & \(+\)0.034 \(\pm\) 0.019 & uniform (marginal) \\
Wanda, kept 0.8 & \best{8.31 \(\pm\) 0.01} & 8.32 \(\pm\) 0.01 & \(+\)0.011 \(\pm\) 0.012 & tie \\
\bottomrule
\end{tabular}
}
\end{table*}

\begin{table*}[!t]
\centering
\small
\caption{\textbf{Seed-robust zero-shot reasoning} (7-task average accuracy\(\uparrow\), n = 5 seeds, mean \(\pm\) std; same cells and protocol as the PPL tables; dense references: LLaMA-2 \(\approx\) 59.6, LLaMA-3 \(\approx\) 60.0). \textit{Takeaway:} every PPL verdict carries over to reasoning accuracy --- the quantization rescue is worth \(+10.8\) points at LLaMA-3 3-bit and \(+3.9\) at 4-bit, OWL's aggressive-sparsity win is \(+2.0\)/\(+2.4\) points over uniform, and DIFW's low-rank edge persists on LLaMA-2 --- with seed noise of only \(\pm\)0.1--1.1 points. In the collapsed low-rank regime (LLaMA-3 SVD) zero-shot cannot discriminate between allocations.}
\label{tab:zs-seeds}
\fitbox
{
\begin{tabular}{llrrrr}
\toprule
\textbf{Cell} & \textbf{Model} & \textbf{uniform} & \textbf{DIFW\textsubscript{Fair} (block)} & \textbf{OWL} & \textbf{HAWQ-style} \\
\midrule
Wanda, kept 0.4 & L2 & 46.3 \(\pm\) 0.1 & 47.0 \(\pm\) 0.2 & \best{48.3 \(\pm\) 0.1} & --- \\
Wanda, kept 0.4 & L3 & 44.8 \(\pm\) 0.2 & 44.6 \(\pm\) 0.1 & \best{47.2 \(\pm\) 0.2} & --- \\
SVD-LLM, kept 0.4 & L2 & 30.2 \(\pm\) 0.2 & \best{30.6 \(\pm\) 0.4} & --- & --- \\
SVD-LLM, kept 0.6 & L2 & 35.2 \(\pm\) 0.2 & \best{35.8 \(\pm\) 0.2} & --- & --- \\
SVD-LLM, kept 0.4 / 0.6 & L3 & 29.7 / 30.9 & 29.7 / 30.8 & --- & --- \\
GPTQ, 3-bit & L2 & 46.0 \(\pm\) 0.8 & 46.6 \(\pm\) 1.0 & --- & \best{47.2 \(\pm\) 1.1} \\
GPTQ, 4-bit & L2 & 53.8 \(\pm\) 0.4 & 53.7 \(\pm\) 0.2 & --- & \best{53.9 \(\pm\) 0.2} \\
GPTQ, 3-bit & L3 & 34.6 \(\pm\) 0.5 & 40.3 \(\pm\) 0.9 & --- & \best{45.4 \(\pm\) 1.1} \\
GPTQ, 4-bit & L3 & 53.8 \(\pm\) 1.9 & 56.4 \(\pm\) 0.5 & --- & \best{57.7 \(\pm\) 0.2} \\
\bottomrule
\end{tabular}
}
\end{table*}

\section{Artifacts and Licenses}
\label{app:licenses}
All models, datasets, and codebases are used for their intended research purpose under their original research or open-source licenses: LLaMA-2 and LLaMA-3 under the Meta Llama community licenses, and BERT, SVD-LLM, Wanda, GPTQ, the allocation baselines, and the lm-evaluation-harness under Apache-2.0 or MIT licenses; WikiText-2 and GLUE are standard public research corpora.

\section{Computational Resources}
\label{app:compute}
All LLM experiments ran on three NVIDIA L40 GPUs (48\,GB each); the BERT ablations used a single GPU.
One compress-and-evaluate configuration takes roughly 20--60 GPU-minutes depending on the base compressor and evaluation suite; the complete experimental program (the \(n=5\) multi-seed grid, the SOTA-baseline runs, and the multi-seed zero-shot suite) totals approximately 40 GPU-days.
The proposed allocation itself is negligible in this budget: one backward pass over the calibration data plus seconds for the closed-form rule or the greedy knapsack.

\end{document}